\RequirePackage{fix-cm}
\documentclass[10pt,twocolumn,letterpaper]{article}

 \usepackage[pagenumbers]{cvpr} 

\usepackage{multirow}
\usepackage{pifont} 
\usepackage{graphicx}
\usepackage{needspace}
\usepackage{array} 

\usepackage{microtype}

\usepackage{cuted}

\usepackage{stfloats}
\usepackage{capt-of}
\usepackage{placeins}
\usepackage{float}

\newcommand{\embTableBefore}{\par\vspace{4pt}\noindent}
\newcommand{\embTableAfter}{\par\vspace{5pt}}

\providecommand{\nolinenumbers}{}
\providecommand{\linenumbers}{}

\definecolor{cvprblue}{rgb}{0.21,0.49,0.74}
\usepackage[pagebackref,breaklinks,colorlinks,
allcolors=cvprblue]{hyperref}

\def\confName{CVPR}
\def\confYear{2027}

\title{Embodied-VGGT: Preserving Reliable Local
Geometry for Long-Sequence Reconstruction}

\author{Yiheng Han, Ziang Chen, Zhuyong Fu, Yiqing Yin, Nan Ma*}

\begin{document}

\maketitle
\nolinenumbers



\begin{strip}
\centering

\includegraphics[
    width=\textwidth,
    trim=0mm 4.8mm 0mm 0mm,
    clip
]{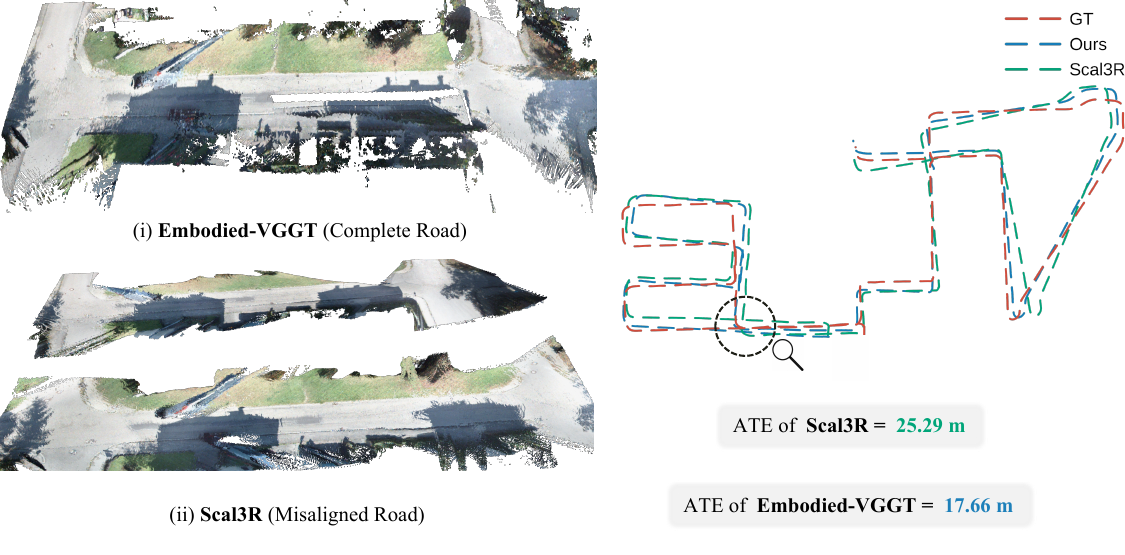}
\captionsetup{hypcap=false}

\captionof{figure}{
Long-sequence 3D reconstruction on KITTI Odometry~\cite{Geiger2012CVPR}.
We highlight a road segment traversed in opposite directions
at different times in the sequence.
\textbf{Embodied-VGGT} reconstructs these
repeated observations as a single coherent road segment,
whereas Scal3R produces visibly misaligned copies.
Over the full 3.22\,km KITTI 08 sequence,
Embodied-VGGT achieves an ATE of 17.66\,m.
}

\label{fig:teaser}
\end{strip}

\linenumbers

\begin{abstract}

Long sequence reconstruction must correct accumulated drift while
preserving locally consistent geometry. In robotics and autonomous
driving, embodied acquisition provides recurring platform-environment
geometry and repeated observations of scene structure and platform
motion.
We present Embodied-VGGT, which exploits these cues to establish
reliable local relationships and selectively preserve them during
global refinement. Memory-Augmented Grounded Chunks (MAGC) builds on Scal3R's global context memory and uses relative scale references derived from predicted ground geometry to regularize test-time memory updates,
encouraging cross-chunk consistency while limiting changes
to reliable local geometry.
Superpoint Alignment (SA) and Cross-Loop Optimization (CL)
jointly address the robust establishment and preservation
of geometric relationships.
SA balances ground and non-ground evidence across spatial regions
to estimate inter-chunk translation. CL uses geometric and motion
consistency across interleaved windows to select coherent chunk
groups, each receiving a shared $\mathrm{Sim}(3)$ correction that
preserves its internal relative transformations during global
optimization. Experiments on five public driving benchmarks demonstrate consistent and substantial improvements in long sequence reconstruction. On KITTI Odometry, Embodied-VGGT achieves a mean absolute trajectory error of 7.30 m and a Chamfer distance of 2.67 m, reducing these errors by 46.8\% and 23.3\% compared with the state-of-the-art baseline under same evaluation protocol. To facilitate reproducibility, we released our code: https://github.com/Ang-Sheep/Embodied-VGGT.

\end{abstract}

\section{Introduction}
\label{sec:intro}

Large-scale 3D reconstruction from long RGB sequences is
fundamental to autonomous driving and robotic mapping. Visual
geometry foundation models such as VGGT~\cite{Wang_2025_CVPR}
predict camera poses and dense scene geometry directly from images,
but extending their predictions to long sequences remains difficult.
Local reconstructions must be registered across changing viewpoints,
while scale and registration errors can accumulate over
substantial travel distances. Global pose refinement may further
compromise otherwise coherent local geometry. The challenge is
therefore not only to connect local reconstructions, but also to
determine which relationships should be preserved during global
adjustment.

Continuous platform-based acquisition offers additional structure
beyond generic multi-view correspondence. Unlike arbitrary image
collections, robotic and driving sequences are captured by a sensing
platform moving through the environment. For ground-based platforms,
approximately stable camera-to-ground relationships provide
relative-scale cues without a known installation height,
as motivated by ground-plane scale estimation in monocular
driving~\cite{Song_2014_CVPR,Zhou_2020_TITS}. Overlapping
windows also reconstruct the same physical frames, yielding repeated
observations of scene geometry and platform motion that can be
cross-checked. The former provides a reference for relating
local scales, while the latter offers evidence for guiding memory
adaptation and assessing inter-chunk consistency. We refer to the
explicit use of these platform--environment geometric and motion
regularities as \emph{embodied reconstruction}.

Existing methods address complementary aspects of this problem.
VGGT-Long~\cite{deng2025vggtlongchunkitloop} combines overlapping
chunk reconstruction with loop closure and global pose refinement.
Scal3R~\cite{Xie_2026_CVPR} shares contextual information across
chunks through neural global context memory, while
VGGT-Align~\cite{zhang2026vggt} uses scene-derived geometric
invariants for cross-chunk scale and TALO~\cite{Zhang_2026_CVPR}
models spatially varying alignment discrepancies. Together, these
approaches demonstrate the value of contextual information and
geometric constraints, yet leave open a complementary question:
how can acquisition-induced cues guide memory adaptation and robust
local registration, while selectively preserving reliable
relationships during global refinement?

This question involves three linked considerations. First,
contextual memory sharing does not by itself ensure geometric
consistency: memory updates should reconcile repeated observations
while limiting changes to reliable local structure. Second,
dense registration can overrepresent spatially concentrated
observations, especially dominant ground surfaces, leaving other
structures underweighted. Third, global refinement must reconcile
adjacent and loop constraints: independent pose corrections can
distort locally coherent relations, whereas preserving every
relation risks retaining erroneous geometry.
The goal is to establish well-supported inter-chunk relations
and preserve only those local relationships consistent across
repeated observations.

We present \textbf{Embodied-VGGT}, an embodied reconstruction
framework that uses platform and environment geometry to guide
memory adaptation, establish reliable local relationships, and
selectively preserve them during global refinement.
\textbf{Memory-Augmented Grounded Chunks (MAGC)} builds on
Scal3R's global context memory and uses relative scale references
derived from predicted ground geometry to regularize test time
memory updates, encouraging cross chunk consistency while limiting
changes to reliable local geometry. This connects contextual
adaptation with explicit geometric grounding, providing a
consistent basis for composing neighboring chunks without requiring
externally provided camera poses or a known installation height.

We further develop a joint strategy for robust geometric relationship
establishment and preservation through \textbf{Superpoint Alignment
(SA)} and \textbf{Cross-Loop Optimization (CL)}.
\textbf{SA} balances spatially structured ground and non-ground
evidence from shared physical frames to estimate inter-chunk
translations at the established scale.
\textbf{CL} converts cross-window geometric and motion consistency
into shared corrections for selected coherent chunk groups.
Each group shares a $\mathrm{Sim}(3)$ correction during pose-graph
optimization, preserving its internal relative transformations
while allowing collective adjustment under adjacent and loop
constraints. The two modules work together to connect robust local
estimation with selective geometry preservation:
\textbf{SA} establishes well-supported relationships, while
\textbf{CL} retains selected relationships as global constraints
are reconciled. Their joint design uses embodied geometric evidence
to guide both local registration and global refinement.

We evaluate Embodied-VGGT on five public driving benchmarks,
including KITTI Odometry~\cite{Geiger2012CVPR},
Virtual KITTI 1~\cite{Gaidon_2016_CVPR},
nuScenes mini~\cite{Caesar_2020_CVPR},
Waymo Open Dataset~\cite{Sun_2020_CVPR},
and Virtual KITTI 2~\cite{Cabon_2020_VKITTI2}.
Results on Waymo Open Dataset and Virtual KITTI 2
are presented in the supplementary material.
Experimental results \mbox{demonstrate} substantial improvements
in 3D reconstruction quality and camera trajectory accuracy over
state-of-the-art chunk-based reconstruction baselines. Our main contributions are summarized as follows:
\begin{itemize}
    \item We introduce Embodied-VGGT, a reconstruction framework that uses platform–environment geometry and repeated observations to guide local prediction and global refinement. Our \textbf{MAGC} module conditions inherited context memory adaptation on ground referenced geometric relations and establishes inter-chunk relative scales without externally provided camera poses.

    \item We develop a joint strategy for robust geometric
    relationship establishment and preservation.
    \textbf{SA} balances spatially structured ground and non-ground
    evidence for translation estimation, while \textbf{CL} converts
    cross-window consistency into shared group corrections that
    preserve selected local geometry during global refinement.

    \item We substantially outperform state-of-the-art methods on KITTI Odometry, Virtual KITTI~1 and~2, nuScenes mini, and Waymo across all evaluated accuracy metrics except RRE, while maintaining practical computational efficiency.
\end{itemize}

\section{Related Work}
\label{sec:related_work}

\noindent\textbf{Long Sequence Visual Geometry.}
Classical structure-from-motion and visual SLAM recover camera motion and scene structure through geometric estimation and global optimization~\cite{Schonberger_2016_CVPR,Campos_2021_TRO,Engel_2014_ECCV}. Learned visual odometry~\cite{Teed_2021_NeurIPS,Teed_2023_NeurIPS} and dense fusion systems~\cite{Whelan_2015_RSS,Dai_2017_TOG} offer complementary tracking and mapping paradigms. Recent feed-forward models, including DUSt3R~\cite{Wang_2024_CVPR}, MASt3R~\cite{Leroy_2024_ECCV}, and VGGT~\cite{Wang_2025_CVPR}, instead infer dense geometry and camera parameters directly from images. Scaling these predictions to extended sequences raises two related challenges: maintaining information across observations and integrating reconstructions produced in different coordinate systems. Recurrent or streaming approaches such as CUT3R~\cite{Wang_CUT3R_2025_CVPR} and LongStream~\cite{Cheng_LongStream_2026_CVPR} propagate temporal information during prediction. VGG-T$^3$~\cite{Elflein_2026_CVPR} studies scalable offline feed-forward inference, while VGGT-Long~\cite{deng2025vggtlongchunkitloop} combines overlapping chunk reconstruction with loop closure and global alignment. Scal3R~\cite{Xie_2026_CVPR} further introduces global-context memory to exchange information across chunks. Although shared context supports local predictions, constructing reliable cross-chunk similarity relations remains a separate geometric problem. We build on this memory-augmented setting and study how those relations can be established and consistently refined.

\noindent\textbf{Geometric Grounding and Structured Alignment.}
Beyond visual correspondence, the geometry of a moving sensing platform can provide recurring constraints on reconstruction. Ground-plane and camera-height relationships have long been used to recover scale in monocular driving~\cite{Song_2014_CVPR,Song_2016_TPAMI,Zhou_2020_TITS}. More recent work, including VGGT-Align~\cite{zhang2026vggt} and LASER~\cite{Ding_LASER_2026_CVPR}, investigates scene-derived scale constraints and alignment across temporal reconstructions. These studies establish geometric grounding as a useful complement to appearance-based registration. A separate consideration is the distribution of alignment evidence: dense correspondences can be concentrated on a limited number of surfaces, while geometrically distinct regions receive comparatively little influence. Robust registration draws on sample consensus~\cite{Fischler_1981_CACM}, global alignment~\cite{Zhou_2016_ECCV}, and certifiable estimation~\cite{Yang_2021_TRO}. Learned matching and overlap-aware registration include SuperGlue~\cite{Sarlin_2020_CVPR}, RPM-Net~\cite{Yew_2020_CVPR}, Predator~\cite{Huang_2021_CVPR}, PointDSC~\cite{Bai_2021_CVPR}, and GeoTransformer~\cite{Qin_2022_CVPR}. SLIC~\cite{Achanta_2012_PAMI} and superpoint partitions~\cite{Landrieu_2018_CVPR,Landrieu_2019_CVPR} motivate region-level organization. In long driving sequences, dominant ground surfaces make this distribution particularly relevant. Overlapping chunks reconstruct the same physical frames, enabling a common geometric partition and hierarchical translation consensus across regions, frames, and ground/non-ground structures under a grounded relative scale.

\noindent\textbf{Global Consistency and Local Structure.}
Loop closure and pose-graph optimization are established tools for reducing accumulated drift~\cite{Mur_Artal_2015_TRO,Kuemmerle_2011_ICRA,Grisetti_2010_ICRA}. Robust graph optimization can also downweight inconsistent closures~\cite{Suenderhauf_2012_IROS}. For large-scale neural reconstruction, VGGT-Long~\cite{deng2025vggtlongchunkitloop} globally optimizes chunk poses, while MERG3R~\cite{Cheng_MERG3R_2026_CVPR} refines integrated reconstructions through global geometric optimization. TALO~\cite{Zhang_2026_CVPR} accommodates spatially varying discrepancies using flexible deformation. These approaches highlight different ways to reconcile local estimates with global constraints, but global adjustment also determines how much of the initial local configuration is allowed to change. We consider this choice through the repeated geometry and motion observations induced by continuous acquisition. Their cross-window agreement guides the selection of chunk groups that share a similarity correction, allowing global adjustment while preserving the selected groups' internal relative transformations.



\begin{figure*}[t]
    \centering

\includegraphics[
    width=\textwidth
]{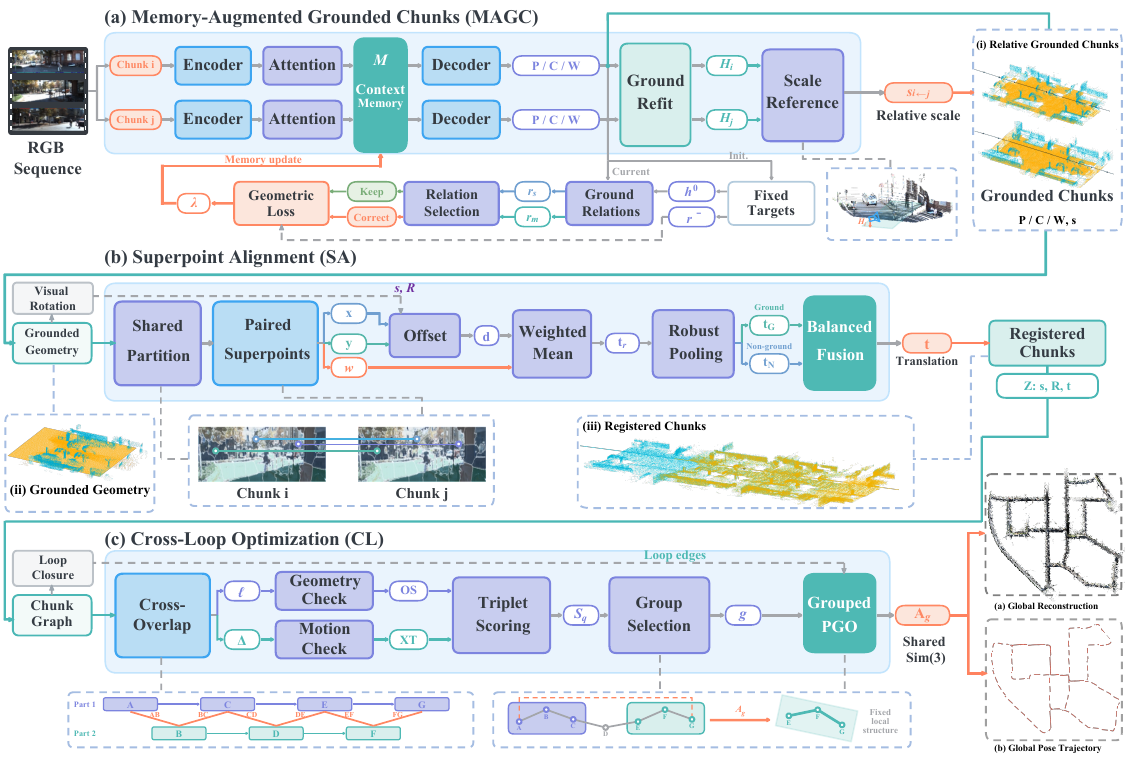}

\captionsetup{skip=3pt}

    \caption{\textbf{Overview of Embodied-VGGT.}
    (a) MAGC regularizes test-time memory adaptation using
ground-referenced geometric relations and establishes
inter-chunk relative-scale references.
(b) SA uses shared spatial partitions to balance ground
and non-ground evidence for translation estimation under
the established scale and retained visual rotation.
(c) CL evaluates geometric and motion consistency across
interleaved windows to select coherent chunk groups,
each sharing a $\mathrm{Sim}(3)$ correction that preserves
internal relative transformations during global pose-graph
optimization.
The refined poses and geometry yield the final
3D reconstruction and camera trajectory.
    }

    \label{fig:overview}
\end{figure*}

\begingroup

\setlength{\abovedisplayskip}{5pt}
\setlength{\abovedisplayshortskip}{3pt}

\setlength{\belowdisplayskip}{5pt}
\setlength{\belowdisplayshortskip}{3pt}

\setlength{\jot}{4pt}

\Needspace{6\baselineskip}

\section{Method}
\label{sec:method}

\subsection{Overview}
\label{sec:method_overview}

Given an ordered RGB sequence $\mathcal I=\{I_f\}_{f=0}^{T-1}$,
we follow Scal3R~\cite{Xie_2026_CVPR} to reconstruct overlapping
chunks $\{\mathcal C_i\}_{i=0}^{N-1}$. Each chunk has an independent
similarity gauge, with corresponding points related by
\begin{equation}
\label{eq:adjacent_sim3}
\mathbf y_p=s_{i\leftarrow j}\mathbf R_{i\leftarrow j}\mathbf x_p
+\mathbf t_{i\leftarrow j}.
\end{equation}
The transform $\mathbf Z_{i\leftarrow j}\in\mathrm{Sim}(3)$ maps
chunk $j$ to $i$~\cite{Umeyama_1991_TPAMI}; $\mathbf C_f^i$
maps camera $f$ to chunk $i$, and $\mathbf G_i$ maps chunk $i$
to the global frame. As shown in Figure~\ref{fig:overview},
MAGC geometrically conditions memory adaptation and relative
scale, SA estimates translation from balanced spatial evidence,
and CL preserves selected local relationships during global refinement.

\subsection{Memory-Augmented Grounded Chunks}
\label{sec:magc}

MAGC combines Scal3R's memory augmentation (MA) with our ground
conditioning (GC), which regularizes memory adaptation and
establishes inter-chunk scales. MA produces initial point maps,
camera poses, and confidences,
$(\mathbf P_0^i,\mathbf C_0^i,\mathbf W_0^i)
=\mathcal F_\theta(\mathcal I_{\mathcal C_i};\mathcal M_{\mathrm{ctx}})$.
GC refines selected runtime memory weights while keeping
pretrained parameters $\theta$ frozen.

From these predictions, we fit a confidence-supported ground
plane $\pi_{i,0}=(\mathbf n_{i,0},b_{i,0})$ with a unit normal
oriented toward predicted camera-up. A shared frame $f$ provides
the reference
\begin{equation}
\label{eq:magc_frame_reference}
h_{if}^{0}=\mathbf n_{i,0}^{\top}\mathbf c_{f,0}^{i}+b_{i,0}>0.
\end{equation}
Repeated predictions observe the same platform configuration,
providing relative length units without a known installation height.
For matching pixel pairs $(p,q)$ and frame pairs $(u,v)$ in an
overlap, we define shape and motion relations
\begin{gather}
r_s^i(f,p,q;\mathcal M)
=\log\frac{\|\mathbf x_{fp}^i(\mathcal M)
-\mathbf x_{fq}^i(\mathcal M)\|_2}{h_{if}^{0}},
\label{eq:magc_relations}\\
\mathbf r_m^i(u,v;\mathcal M)
=\frac{\mathbf R_u^i(\mathcal M)^\top
\bigl(\mathbf c_v^i(\mathcal M)-\mathbf c_u^i(\mathcal M)\bigr)}
{h_{iu}^{0}}.
\label{eq:magc_motion_relation}
\end{gather}
Here $\mathbf x_{fp}^i$ is a point-map entry and
$(\mathbf R_u^i,\mathbf c_u^i)$ is a camera-to-chunk pose.
Distances and displacements in the shared camera frame are
comparable across chunks without coordinate alignment.
Invalid references and degenerate point pairs are excluded.

Among well-supported relations, small initial disagreements
form a preservation set $\mathcal K$, and moderate disagreements
form a correction set $\mathcal R$; unreliable or severely
inconsistent observations are discarded. The initial references and consensus targets remain fixed throughout refinement. This prevents changes in the normalization references from absorbing the geometric discrepancies being optimized and provides stable targets for both preserving initially consistent relations and reconciling moderate disagreements. Each target $\bar r_{e,t}$
is the mean of the two initial predictions weighted by confidence
and ground-fit quality. References, targets, samples, weights,
and set assignments remain fixed during refinement. For relation
type $t\in\{s,m\}$, we use
\begin{equation}
\label{eq:magc_relation_loss}
\mathcal L_t(\mathcal S;\mathcal M)
=\operatorname{Avg}_{e\in\mathcal S_t}\frac12
\sum_{\ell\in\{i_e,j_e\}}
\rho\!\left(r_{e,t}^{\ell}(\mathcal M)-\bar r_{e,t}\right),
\end{equation}
where $i_e,j_e$ are the observing chunks and $\rho$ is a Huber
penalty, summed componentwise for vectors. Averaging proceeds
within each frame or frame pair and then across groups;
empty sets contribute zero.
The geometric loss is
\begin{equation}
\label{eq:magc_objective}
\mathcal L_{\mathrm{geo}}
=\sum_{t\in\{s,m\}}\beta_t\left[
\mathcal L_t(\mathcal R;\mathcal M)
+\beta_{\mathrm{keep}}\mathcal L_t(\mathcal K;\mathcal M)\right].
\end{equation}
We weight it as $\mathcal L_{\mathrm{add}}=\lambda\mathcal L_{\mathrm{geo}}$,
where $\beta_s=1$, $\beta_m$ and $\beta_{\mathrm{keep}}$ balance
the terms, and $\lambda$ controls regularization strength.
Starting from $\mathcal M^{(0)}=\mathcal M_{\mathrm{ctx}}$,
we back-propagate through the predicted geometry for $K$ steps:
\begin{equation}
\label{eq:magc_memory_update}
\mathcal M^{(k+1)}=\mathcal M^{(k)}
-\lambda\mathcal D\!\left(
\nabla_{\mathcal M}\mathcal L_{\mathrm{geo}}(\mathcal M^{(k)})\right),
\end{equation}
where $\mathcal D$ applies a fixed step size and update-norm clipping.

GC then decodes predictions using $\mathcal M^{(K)}$, refits
valid ground planes $\pi_i=(\mathbf n_i,b_i)$, and establishes
chunk references and relative scales:
\begin{gather}
H_i=\operatorname*{median}_{f\in\mathcal F_i}
(\mathbf n_i^\top\mathbf c_f^i+b_i)>0,
\label{eq:magc_height}\\
s_{i\leftarrow j}^{\mathrm{MAGC}}=H_i/H_j.
\label{eq:magc_scale}
\end{gather}
Here $\mathcal F_i$ contains the reference-estimation frames.
Unlike the fixed $h_{if}^{0}$, $H_i$ reflects the refined geometry.
The scale ratio assumes locally stable platform--ground geometry
and does not recover absolute metric scale. Edges without valid
references retain visual similarity registration. Reference
selection, scale derivation, and refinement settings are given
in the supplementary material.

MAGC thus provides geometrically regularized chunk predictions
and relative-scale references for subsequent inter-chunk
alignment, without requiring externally provided camera poses.

\subsection{Superpoint Alignment}
\label{sec:superpoint_alignment}

Overlapping chunks reconstruct the same physical frames, so shared
pixels provide direct 3D correspondences. Pooling all correspondences
together can give densely sampled regions disproportionate influence.
SA organizes correspondences into common spatial regions and
aggregates translation evidence across regions and frames before
balancing the ground and non-ground estimates. With scale supplied
by MAGC and rotation retained from robust visual registration, we
first compute the pointwise translation offsets:
\begin{equation}
\label{eq:sp_translation_observation}
\mathbf d_p=\mathbf y_p-s_{i\leftarrow j}^{\mathrm{MAGC}}
\mathbf R_{i\leftarrow j}^{\mathrm{vis}}\mathbf x_p.
\end{equation}
For each shared frame, we construct a single deterministic SLIC-inspired partition ~\cite{Achanta_2012_PAMI} using RGB, pixel proximity, normalized depth, surface geometry, and symmetric camera-coordinate cues from the two predictions. This partition assigns identical region labels to corresponding pixels in both point maps, defining pairs of 3D superpoints with shared pixel membership. Frame-local planes, checked
against chunk references, classify correspondences as ground
($\mathrm G$) or non-ground ($\mathrm N$) only when both
predictions agree; uncertain observations are excluded.

Let $\Omega_{f,r,c}$ denote valid correspondences in frame $f$, shared region $r$, and category $c$. Confidence-weighted regional translation estimates are aggregated using the geometric median $\operatorname{gmed}$ over regions and then frames, with equal ground/non-ground weights when both categories have sufficient support:

\begin{gather}
\mathbf t_{f,r,c}
=\frac{\sum_{p\in\Omega_{f,r,c}}w_p\mathbf d_p}
{\sum_{p\in\Omega_{f,r,c}}w_p},
\label{eq:sp_region_estimator}\\
\mathbf t_{f,c}=\operatorname{gmed}_r\{\mathbf t_{f,r,c}\},
\label{eq:sp_two_level}\\
\mathbf t_c=\operatorname{gmed}_f\{\mathbf t_{f,c}\},
\label{eq:sp_category_estimator}\\
\mathbf t_{i\leftarrow j}^{\mathrm{SP}}
=\tfrac12(\mathbf t_{\mathrm G}+\mathbf t_{\mathrm N}).
\label{eq:sp_final_translation}
\end{gather}
Each region contributes one representative regardless of pixel
coverage, preventing dense ground observations from dominating.
With insufficient category support, SA uses an all-correspondence
Huber translation estimate. The resulting edge is
$\mathbf Z_{i\leftarrow j}=(s_{i\leftarrow j}^{\mathrm{MAGC}},
\mathbf R_{i\leftarrow j}^{\mathrm{vis}},
\mathbf t_{i\leftarrow j}^{\mathrm{SP}})$.
Partition construction and support tests are specified in the supplement.

\begin{table*}[!b]
\centering
\begingroup
\setlength{\tabcolsep}{1.65pt}
\renewcommand{\arraystretch}{1.13}
\fontsize{8.0}{9.4}\selectfont
\newcommand{\failedrun}[1]{{\color{black!55}\textit{#1}}}
\resizebox{\textwidth}{!}{%
\begin{tabular}{@{\hspace{1.6pt}}l !{\hspace{1.6pt}\vrule width .28pt\hspace{1.6pt}} *{11}{c} !{\hspace{1.6pt}\vrule width .28pt\hspace{1.6pt}} c !{\hspace{1.6pt}\vrule width .28pt\hspace{1.6pt}} c c c@{\hspace{2pt}}}
\toprule
\textbf{Methods}
 & \multicolumn{11}{c !{\hspace{1.6pt}\vrule width .28pt\hspace{1.6pt}}}{\textbf{KITTI Odometry -- per-sequence ATE (m) $\downarrow$}}
 & \multicolumn{4}{c}{\textbf{Summary}} \\
\cmidrule(lr){2-12}\cmidrule(l){13-16}
 & \textbf{00} & \textbf{01} & \textbf{02} & \textbf{03} & \textbf{04} & \textbf{05}
 & \textbf{06} & \textbf{07} & \textbf{08} & \textbf{09} & \textbf{10}
 & \textbf{Valid} & \textbf{ATE} & \textbf{RTE} & \textbf{RRE} \\
\textit{seq. frames} & 4541 & 1101 & 4661 & 801 & 271 & 2761
 & 1101 & 1101 & 4071 & 1591 & 1201 & -- & -- & -- & -- \\
\midrule
DROID-SLAM~\cite{Teed_2021_NeurIPS} & 110.13 & 90.85 & 138.87 & 3.20 & 1.47 & 73.65 & 61.48 & 18.61 & 104.22 & 89.69 & 22.24 &11/11& 64.95 & 25.95 & \textbf{0.50} \\
MASt3R-SLAM~\cite{Murai_2025_CVPR} & 189.58 & 529.52 & 282.08 & 22.40 & 88.99 & 155.32 & 77.10 & 81.10 & 263.75 & 199.17 & 152.10 &11/11& 185.56 & 82.15 & 28.74 \\
\addlinespace[1pt]
\cmidrule(lr){1-16}
CUT3R~\cite{Wang_CUT3R_2025_CVPR} & 191.04 & 626.75 & 299.78 & 153.83 & 25.61 & 152.65 & 132.23 & 74.86 & 230.75 & 205.99 & 181.42 &11/11& 206.81 & 96.75 & 37.48 \\
Fast3R~\cite{Yang_Fast3R_2025_CVPR} & \failedrun{OOM} & 722.01 & \failedrun{OOM} & \failedrun{PF} & 112.97 & \failedrun{OOM} & 137.45 & 89.59 & \failedrun{OOM} & \failedrun{OOM} & 211.43 &5/11& -- & -- & -- \\
FastVGGT~\cite{Shen_2026_ICLR} & \failedrun{OOM} & 638.61 & \failedrun{OOM} & 21.88 & 9.99 & \failedrun{OOM} & 41.37 & 51.84 & \failedrun{OOM} & 201.14 & 197.13 &7/11& -- & -- & -- \\
STream3R~\cite{Lan_STream3R_2026_ICLR} & \failedrun{OOM} & \failedrun{OOM} & \failedrun{OOM} & \failedrun{OOM} & 71.05 & \failedrun{OOM} & \failedrun{OOM} & \failedrun{OOM} & \failedrun{OOM} & \failedrun{OOM} & \failedrun{OOM} &1/11& -- & -- & -- \\
TTT3R~\cite{Chen_TTT3R_2026_ICLR} & 178.72 & 533.66 & 254.08 & 109.53 & 10.51 & 146.29 & 131.99 & 63.49 & 259.38 & 188.07 & 123.89 &11/11& 181.78 & 84.86 & 30.34 \\
\addlinespace[1pt]
\cmidrule(lr){1-16}
VGGT-Long~\cite{deng2025vggtlongchunkitloop} & 8.64 & 58.37 & 52.80 & 8.82 & 4.23 & 9.85 & 4.67 & 2.66 & 73.12 & 31.85 & 27.71 &11/11& 25.70 & 12.41 & 3.15 \\
SwiftVGGT~\cite{Lee_SwiftVGGT_2025_arxiv} & 7.79 & 146.38 & 50.32 & 13.44 & 5.79 & 11.04 & 8.40 & 2.80 & 79.28 & 28.85 & 30.87 &11/11& 35.00 & 16.14 & 3.42 \\
Scal3R~\cite{Xie_2026_CVPR} & 4.18 & 50.47 & 37.77 & 5.22 & 1.05 & 3.84 & 3.37 & 1.36 & 25.29 & \textbf{12.43} & 5.96 &11/11& 13.72 & 5.88 & 1.65 \\
VGGT-Align~\cite{zhang2026vggt} & 7.68 & 65.26 & 46.99 & 5.04 & 1.73 & 4.92 & 4.23 & 4.47 & 44.10 & 23.24 & 11.77 &11/11& 19.95 & 9.61 & 3.16 \\
TALO~\cite{Zhang_2026_CVPR} & 54.17 & 76.87 & 108.89 & 6.05 & 3.94 & 48.88 & 31.70 & 13.03 & 52.18 & 51.91 & 21.93 &11/11& 42.69 & 14.53 & 3.51 \\
\midrule
\textbf{Embodied-VGGT (Ours)} & \textbf{3.80} & \textbf{19.19} & \textbf{14.02} & \textbf{1.79} & \textbf{1.02} & \textbf{2.49} & \textbf{1.29} & \textbf{1.20} & \textbf{17.66} & 12.57 & \textbf{5.25} &11/11& \textbf{7.30} & \textbf{3.97} & 1.81 \\
\bottomrule
\end{tabular}%
}
\endgroup
\captionof{table}{\textbf{Camera pose accuracy on KITTI Odometry.}
Per-sequence absolute trajectory error (ATE, m), mean ATE (m),
relative translation error (RTE, m/100\,m), and relative rotation error
(RRE, deg/100\,m).
Mean errors are reported only for methods completing all 11 sequences. Lower is better.
Bold indicates the lowest error; 
-- denotes no complete-run mean;
\textit{OOM}: out of memory; \textit{PF}: pose-estimation
failure.}
\label{tab:pose_kitti_all}
\end{table*}

\subsection{Cross-Loop Optimization}
\label{sec:crossloop}

Adjacent relations initialize $\mathbf G_0^0=\mathbf I$,
$\mathbf G_{k+1}^0=\mathbf G_k^0\mathbf Z_{k\leftarrow k+1}$.
With 50\% overlap, alternating chunks form two interleaved sequences, each containing nonoverlapping windows. CL checks candidate triplets $q=(A,B,C)$, where $B$ overlaps both $A$ and $C$, using repeated geometry and motion in the initialized global frame.

Overlap-scale inconsistency (OS) compares matched
point-pair distances $\ell^a,\ell^b$ from the same physical frame:
\begin{gather}
e_{a,b,f,u,v}^{\mathrm{OS}}
=\left|\log\frac{\ell_{f,u,v}^a}{\ell_{f,u,v}^b}\right|,
\label{eq:cl_os_pair}\\
\mathrm{OS}_{a,b}
=Q_{.90,f}\!\left[Q_{.90,(u,v)}e_{a,b,f,u,v}^{\mathrm{OS}}\right],
\label{eq:cl_os_aggregate}\\
\mathrm{OS}(q)
=\max\{\mathrm{OS}_{A,B},\mathrm{OS}_{B,C}\}.
\label{eq:cl_os_triplet}
\end{gather}
here $Q$ denotes a quantile, applied first within frames and
then across frames. Cross-window motion inconsistency (XT) compares motion spanning the two overlaps, $u\in A\cap B$ and $v\in B\cap C$, as predicted jointly by $A,C$ and independently by $B$:\begin{gather}
\boldsymbol\Delta_{AC}
=(\mathbf R_u^A)^\top(\mathbf c_v^C-\mathbf c_u^A),
\label{eq:cl_motion}\\
\boldsymbol\Delta_B
=(\mathbf R_u^B)^\top(\mathbf c_v^B-\mathbf c_u^B),
\label{eq:cl_motion_center}\\
\mathrm{XT}(q)
=Q_{.75,h}\!\left[
\frac{\|\boldsymbol\Delta_{AC}-\boldsymbol\Delta_B\|_2}
{n_{uv}\ell_B}\right].
\label{eq:cl_xt_span}
\end{gather}
here $h$ indexes spans, $n_{uv}$ is physical-frame separation,
and $\ell_B$ is a robust camera-step scale. All centers and
orientations above are expressed in the initialized global frame.

For valid triplets, sequence-normalized ranks $u_q,v_q$ of
OS and XT define
\begin{equation}
\label{eq:cl_selection_score}
S(q)=\max\{u_q,v_q\}+\frac{u_q+v_q}{8}.
\end{equation}

We greedily select nonoverlapping low-score triplets from a
fixed-budget eligible pool; remaining chunks are singletons.
Triplets containing both endpoints of a loop edge are excluded
to keep that constraint between groups. Selection reflects
cross-window agreement rather than externally verified accuracy;
eligibility and tie-handling rules are in the supplement.

For each triplet or singleton $g$, we optimize one shared left
correction $\mathbf A_g=\operatorname{Exp}(\boldsymbol\xi_g)$,
$\boldsymbol\xi_g\in\mathfrak{sim}(3)$. With $g(i)$ denoting
chunk $i$'s group, the parameterization preserves relative
transforms whenever $g(i)=g(j)$:
\begin{gather}
\mathbf G_i=\mathbf A_{g(i)}\mathbf G_i^0,
\label{eq:cl_group_parameterization}\\
\mathbf G_i^{-1}\mathbf G_j
=(\mathbf G_i^0)^{-1}\mathbf G_j^0.
\label{eq:cl_relative_invariance}
\end{gather}
We retain the original graph measurements and objective:
\begin{equation}
\label{eq:cl_graph_objective}
\min_{\{\boldsymbol\xi_g\}}\frac{1}{|\mathcal E|}
\sum_{(i,j)\in\mathcal E}
\left\|\operatorname{Log}
(\mathbf M_{ij}\mathbf X_i\mathbf X_j^{-1})\right\|_2^2,
\end{equation}
where $\mathbf X_i=\mathbf G_i^{-1}$ and $\mathbf M_{ij}$ is
the inverse source-to-target measurement
($\mathbf M_{ij}=\mathbf Z_{i\leftarrow j}^{-1}$ for adjacent edges).
Anchoring the first chunk's group removes global gauge freedom.
Global constraints thus adjust each selected group collectively
while preserving its internal relative $\mathrm{Sim}(3)$ transforms.

\endgroup


\flushbottom
\Needspace{5\baselineskip}
\section{Experiments}
\label{sec:experiments}

\subsection{Experimental Setup}
\label{sec:experimental_setup}

We evaluate Embodied-VGGT on KITTI Odometry~\cite{Geiger2012CVPR}
(11 sequences), Virtual KITTI 1 Scene 20~\cite{Gaidon_2016_CVPR}
(six appearance conditions), and nuScenes mini~\cite{Caesar_2020_CVPR}
(ten sequences).
Additional evaluations on Waymo~\cite{Sun_2020_CVPR}
and Virtual KITTI 2~\cite{Cabon_2020_VKITTI2}
are presented in the supplementary material.

For Camera Pose Accuracy, we evaluate long driving trajectories
on KITTI Odometry, shorter trajectories on nuScenes mini,
and six different weather conditions on Virtual KITTI 1 Scene 20.
For 3D Reconstruction Quality, we evaluate point-cloud geometry
on KITTI Odometry.
Additionally we evaluate computational efficiency on KITTI Odometry
to compare the computational costs of different methods.
Comparisons cover visual SLAM, feed-forward, streaming,
and chunk-based approaches.
For chunk-based methods, we use 60-frame windows
with 30-frame overlap, while other methods follow
their native inference protocols.

Experiments use two NVIDIA RTX 5880 Ada GPUs,
an Intel Core Ultra 9 285K CPU, and 192 GiB RAM.

\subsection{Camera Pose Accuracy}
\label{sec:pose_experiments}

We report absolute trajectory error (ATE) after Sim(3) alignment;
for KITTI we also report relative translation error (RTE) and relative
rotation error (RRE) over 100\,m segments.

\begin{figure*}[!t]
    \centering

    \makebox[\textwidth][l]{%
        \hspace*{-4mm}%
        \includegraphics[
            width=1.01\textwidth,
            trim=0pt 8pt 5pt 0pt,
            clip
        ]{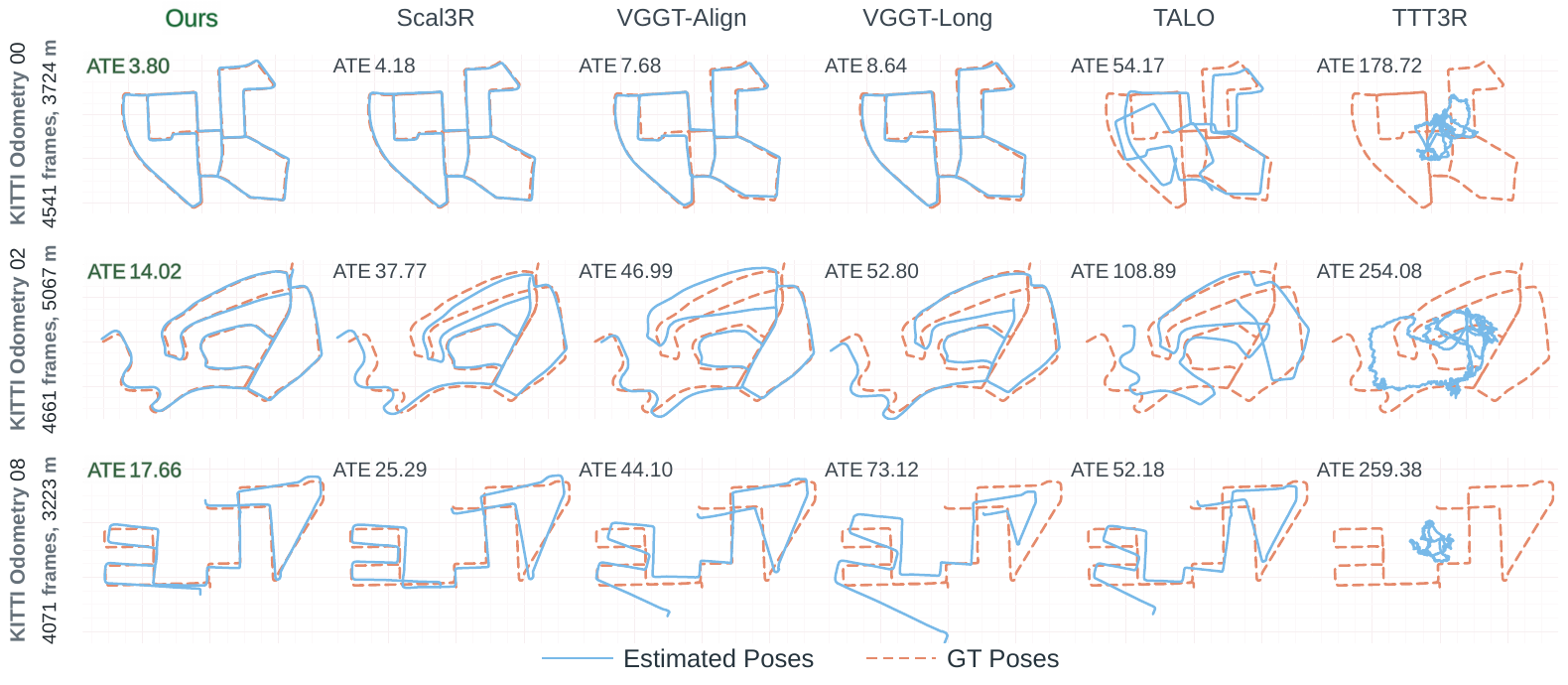}%
    }

    \caption{\textbf{Camera trajectory comparisons on KITTI Odometry.}
    Estimated camera trajectories and ground truth on sequences 00, 02,
    and 08 for Embodied-VGGT and five representative baselines.
    ATE values (m) are displayed in each panel.}

    \label{fig:kitti_trajectories}
\end{figure*}

\noindent\textbf{KITTI Odometry.}
Table~\ref{tab:pose_kitti_all} reports camera pose accuracy on all 11 KITTI sequences. Embodied-VGGT achieves the lowest mean ATE of 7.30 m and mean RTE of 3.97 m/100 m, improving over Scal3R by 46.8\% and 32.5\%, respectively. It obtains the lowest ATE on 10 of the 11 sequences, with particularly large reductions on sequences 01 and 02: from 50.47 to 19.19 m and from 37.77 to 14.02 m. Sequence 09 remains slightly behind Scal3R. The improvements are strongest in trajectory position and relative translation; mean RRE is 1.81 deg/100 m, compared with 1.65 for Scal3R. Fig.~\ref{fig:kitti_trajectories} visualizes representative trajectory comparisons.


\noindent\textbf{Virtual KITTI 1 Scene 20.}
Across the six appearance conditions, Embodied-VGGT obtains the
lowest ATE in each condition (Table~\ref{tab:pose_vkitti_single}).
These variants share the underlying scene and camera path but change
weather or illumination, making them useful for examining visual
appearance sensitivity. Our ATE ranges from 1.59 to 2.61\,m,
with a six-condition mean of 2.05\,m. The corresponding mean is
4.23\,m for Scal3R and 4.31\,m for VGGT-Align.

\begin{table*}[!b]
\centering
\vspace{0pt}
\begingroup
\setlength{\tabcolsep}{6.1pt}
\renewcommand{\arraystretch}{1.11}
\fontsize{8.5}{10.0}\selectfont
\newcommand{\unscored}[1]{{\color{black!55}\textit{#1}}}
\begin{tabular*}{\textwidth}{@{\hspace{3pt}}l@{\hspace{3pt}\vrule width .28pt\hspace{3pt}\extracolsep{\fill}}cccccc@{\hspace{5pt}\vrule width .28pt\hspace{5pt}}>{\centering\arraybackslash}p{45pt}@{\hspace{10pt}}}
\toprule
\textbf{Methods} & \textbf{Clone} & \textbf{Fog} & \textbf{Morning} & \textbf{Overcast} & \textbf{Rain} & \textbf{Sunset} & \textbf{Mean} \\
\midrule
DROID-SLAM~\cite{Teed_2021_NeurIPS} & 6.42 & 3.10 & 3.42 & 4.02 & 4.83 & 6.20 & 4.66 \\
MASt3R-SLAM~\cite{Murai_2025_CVPR} & 229.27 & \unscored{PF} & 230.34 & 231.06 & 231.14 & 231.06 & -- \\
\midrule
CUT3R~\cite{Wang_CUT3R_2025_CVPR} & 131.15 & 76.49 & 109.02 & 119.06 & 70.92 & 102.30 & 101.48 \\
Fast3R~\cite{Yang_Fast3R_2025_CVPR} & 230.74 & 231.12 & 230.67 & 230.25 & 229.77 & 229.75 & 230.38 \\
FastVGGT~\cite{Shen_2026_ICLR} & 28.69 & 39.95 & 24.37 & 20.32 & 33.27 & 21.82 & 28.07 \\
STream3R~\cite{Lan_STream3R_2026_ICLR} & \unscored{OOM} & \unscored{OOM} & \unscored{OOM} & \unscored{OOM} & \unscored{OOM} & \unscored{OOM} & -- \\
TTT3R~\cite{Chen_TTT3R_2026_ICLR} & 65.02 & 62.78 & 65.00 & 69.55 & 60.31 & 62.37 & 64.17 \\
\midrule
VGGT-Long~\cite{deng2025vggtlongchunkitloop} & 8.91 & 9.59 & 4.86 & 2.49 & 3.31 & 2.89 & 5.34 \\
SwiftVGGT~\cite{Lee_SwiftVGGT_2025_arxiv} & 6.81 & 9.33 & 2.29 & 4.60 & 3.18 & 4.76 & 5.16 \\
Scal3R~\cite{Xie_2026_CVPR} & 5.36 & 5.53 & 3.56 & 4.51 & 3.71 & 2.70 & 4.23 \\
VGGT-Align~\cite{zhang2026vggt} & 3.95 & 6.26 & 3.04 & 4.27 & 4.66 & 3.65 & 4.31 \\
TALO~\cite{Zhang_2026_CVPR} & 19.03 & 19.47 & 18.57 & 19.24 & 21.12 & 19.21 & 19.44 \\
\midrule
\textbf{Embodied-VGGT (Ours)} & \textbf{2.61} & \textbf{1.94} & \textbf{2.20} & \textbf{1.83} & \textbf{2.14} & \textbf{1.59} & \textbf{2.05} \\
\bottomrule
\end{tabular*}
\endgroup
\caption{\textbf{Camera pose accuracy on Virtual KITTI 1 Scene 20.}
ATE (m) over six weather and lighting conditions. Lower is better. 
Bold indicates the lowest error; 
-- denotes no complete-run mean;
\textit{OOM}: out of memory; \textit{PF}: pose-estimation
failure.}
\label{tab:pose_vkitti_single}
\end{table*}

\noindent\textbf{nuScenes mini.}
Embodied-VGGT achieves the best performance with the lowest mean ATE of
0.78\,m across all ten scenes
(Table~\ref{tab:pose_nuscenes_single}).
Compared with Scal3R (1.17\,m) and VGGT-Align
(1.45\,m), Embodied-VGGT demonstrates improved
camera pose accuracy while producing valid predictions
for all ten scenes.

\begin{table}[!t]
\centering
\begingroup
\setlength{\tabcolsep}{3.1pt}
\renewcommand{\arraystretch}{1.0}
\fontsize{8.25}{9.4}\selectfont
\begin{tabular*}{\columnwidth}{@{\extracolsep{\fill}}l!{\hspace{3pt}\vrule width .28pt\hspace{3pt}}cc@{\hspace{6pt}}}
\toprule
\textbf{Methods} & \textbf{Valid} & \textbf{Mean ATE (m)}$\downarrow$ \\
\midrule
CUT3R~\cite{Wang_CUT3R_2025_CVPR} & 10/10 & 8.20 \\
Fast3R~\cite{Yang_Fast3R_2025_CVPR} & 8/10 & -- \\
FastVGGT~\cite{Shen_2026_ICLR} & 10/10 & 2.38 \\
STream3R~\cite{Lan_STream3R_2026_ICLR} & 10/10 & 10.20 \\
TTT3R~\cite{Chen_TTT3R_2026_ICLR} & 10/10 & 3.10 \\
\midrule
VGGT-Long~\cite{deng2025vggtlongchunkitloop} & 10/10 & 1.72 \\
SwiftVGGT~\cite{Lee_SwiftVGGT_2025_arxiv} & 10/10 & 1.55 \\
Scal3R~\cite{Xie_2026_CVPR} & 10/10 & 1.17 \\
VGGT-Align~\cite{zhang2026vggt} & 10/10 & 1.45 \\
TALO~\cite{Zhang_2026_CVPR} & 10/10 & 1.30 \\
\midrule
\textbf{Embodied-VGGT (Ours)} & 10/10 & \textbf{0.78} \\
\bottomrule
\end{tabular*}
\endgroup
\caption{\textbf{Camera pose accuracy on nuScenes mini.}
Mean ATE (m) across ten scenes. Lower is better. Bold
indicates the lowest error;
-- denotes no complete-run mean.}
\label{tab:pose_nuscenes_single}
\end{table}


\subsection{3D Reconstruction Quality}
\label{sec:geometry_results}

Following prior work ~\cite{Xie_2026_CVPR} ~\cite{zhang2026vggt}, we evaluate 3D reconstruction quality on KITTI Odometry using accuracy (Acc.), completeness (Comp.), and Chamfer distance (CD). These metrics compare reconstructed and reference point clouds expressed in a common coordinate system. Acc. measures the mean Euclidean distance from each reconstructed point to its nearest reference point, reflecting geometric accuracy. Comp. measures the reverse distance, from each reference point to its nearest reconstructed point, assessing coverage of the reference geometry. CD is defined as the arithmetic mean of these two directional errors, jointly accounting for accuracy and completeness.

Table~\ref{tab:recon_quality} evaluates reconstructed point clouds on KITTI using a common reference geometry and evaluation protocol. Embodied-VGGT achieves the lowest mean accuracy error, completeness error, and Chamfer distance of 3.80, 1.54, and 2.67 m, respectively, compared with 4.56, 2.41, and 3.48 m for Scal3R. The 23.3\% reduction in Chamfer distance, together with improvements in both directional errors, shows that the trajectory gains are accompanied by better geometric agreement and coverage.

\nolinenumbers
\embTableBefore
\begin{minipage}{\columnwidth}
\centering
\begingroup
\setlength{\tabcolsep}{3.7pt}
\renewcommand{\arraystretch}{0.97}
\fontsize{8.3}{9.2}\selectfont
\begin{tabular*}{\columnwidth}{@{\extracolsep{\fill}}lccc@{\hspace{6pt}}}
\toprule
\textbf{Methods} & \textbf{Acc.}$\downarrow$ & \textbf{Comp.}$\downarrow$ & \textbf{CD}$\downarrow$ \\
\midrule
VGGT-Long~\cite{deng2025vggtlongchunkitloop} & 6.42 & 9.18 & 7.80 \\
SwiftVGGT~\cite{Lee_SwiftVGGT_2025_arxiv} & 8.30 & 10.86 & 9.58 \\
VGGT-Align~\cite{zhang2026vggt} & 5.30 & 6.13 & 5.72 \\
TALO~\cite{Zhang_2026_CVPR} & 7.94 & 11.66 & 9.80 \\
TTT3R~\cite{Chen_TTT3R_2026_ICLR} & 33.92 & 51.90 & 42.91 \\
Scal3R~\cite{Xie_2026_CVPR} & 4.56 & 2.41 & 3.48 \\
\midrule
\textbf{Embodied-VGGT (Ours)} & \textbf{3.80} & \textbf{1.54} & \textbf{2.67} \\
\bottomrule
\end{tabular*}
\endgroup
\captionsetup{hypcap=false}
\captionof{table}{\textbf{3D reconstruction quality on KITTI Odometry.}
Mean Acc., Comp., and CD (m) across 11 sequences.
Lower is better.}
\label{tab:recon_quality}
\end{minipage}
\vspace{-1pt}
\linenumbers
\vspace{-6pt}

\subsection{Computational Efficiency}
\label{sec:resource_efficiency}

\nolinenumbers
\embTableBefore
\begin{minipage}{\columnwidth}
\centering
\begingroup
\setlength{\tabcolsep}{1.0pt}
\renewcommand{\arraystretch}{1.10}
\fontsize{7.7}{9.0}\selectfont

\begin{tabular*}{\columnwidth}
{@{}l!{\hspace{1.5pt}\vrule width .28pt\hspace{1.5pt}}@{\extracolsep{\fill}}cccc@{}}

\toprule
\textbf{Methods} &
\textbf{Run time} &
\textbf{Total Chunk} &
\textbf{Align} &
\textbf{Optim} \\

\midrule

DROID-SLAM~\cite{Teed_2021_NeurIPS}
& 154.6 & -- & -- & -- \\

MASt3R-SLAM~\cite{Murai_2025_CVPR}
& 149.6 & -- & -- & -- \\

\addlinespace[1.4pt]
\midrule

VGGT-Long~\cite{deng2025vggtlongchunkitloop}
& 412.5 & 106.0 & 222.5 & 84.0 \\

SwiftVGGT~\cite{Lee_SwiftVGGT_2025_arxiv}
& \textbf{119.9} & \textbf{74.1}
& \textbf{7.6} & \textbf{38.2} \\

VGGT-Align~\cite{zhang2026vggt}
& 365.5 & 118.7 & 203.0 & 43.8 \\

TALO~\cite{Zhang_2026_CVPR}
& 519.6 & 101.4 & 53.5 & 364.7 \\

Scal3R~\cite{Xie_2026_CVPR}
& 673.6 & 559.4 & 29.0 & 85.2 \\

\midrule

\textbf{Embodied-VGGT (Ours)}
& 749.0 & 556.9 & 39.6 & 152.5 \\

\bottomrule
\end{tabular*}

\endgroup

\captionsetup{hypcap=false}

\captionof{table}{
\textbf{Computational efficiency on KITTI Odometry.}
Mean time (s) over 11 sequences.
Run time denotes total execution time;
Total Chunk is the accumulated chunk-processing time;
Align denotes alignment time;
Optim includes global optimization, data transfer,
and other processing overhead.
-- indicates unavailable measurements.
}

\label{tab:resource_runtime}
\end{minipage}
\embTableAfter
\linenumbers

As shown in Table~\ref{tab:resource_runtime},
compared with Scal3R, the total runtime of Embodied-VGGT increases from 673.6 to 749.0 s (11.2\%), while mean ATE decreases from 13.72 to 7.30 m (46.8\%), providing a favorable trade-off between reconstruction accuracy and computational cost.

\subsection{Ablation Study}
\label{sec:ablation_experiments}

We evaluate 9 configurations of inherited memory augmentation
(MA), our ground conditioning (GC), Superpoint Alignment (SA),
and Cross-Loop Optimization (CL) on KITTI 00--10
(Table~\ref{tab:ablation_factors}).

With SA and CL disabled, adding GC to MA reduces mean ATE
from 13.65\,m to 8.36\,m, a 38.8\% reduction. This supports
the benefit of ground conditioning for memory-augmented
reconstruction.

\begin{table}[!htbp]
\centering
\begingroup
\newcommand{\abYes}{\ding{51}}
\newcommand{\abNo}{\ding{55}}
\setlength{\tabcolsep}{4.6pt}
\renewcommand{\arraystretch}{0.99}
\fontsize{8.45}{10.0}\selectfont
\begin{tabular*}{\columnwidth}
{@{\extracolsep{\fill}}ccccc@{\hspace{6pt}}}
\toprule
\multicolumn{2}{c}{\textbf{MAGC}}
& \multirow{2}{*}{\textbf{SA}}
& \multirow{2}{*}{\textbf{CL}}
& \multirow{2}{*}{\textbf{ATE (m)}$\downarrow$} \\
\cmidrule(lr){1-2}
\textbf{MA} & \textbf{GC} & & & \\
\midrule
\abNo  & \abNo  & \abNo  & \abNo  & 25.70 \\
\abYes & \abNo  & \abNo  & \abNo  & 13.65 \\
\abYes & \abNo  & \abNo  & \abYes & 13.52 \\
\abYes & \abNo  & \abYes & \abNo  & 13.54 \\
\abYes & \abNo  & \abYes & \abYes & 13.42 \\
\abYes & \abYes & \abNo  & \abNo  & 8.36 \\
\abYes & \abYes & \abNo  & \abYes & 8.18 \\
\abYes & \abYes & \abYes & \abNo  & 8.32 \\
\midrule
\abYes & \abYes & \abYes & \abYes & \textbf{7.30} \\
\bottomrule
\end{tabular*}
\endgroup
\caption{\textbf{Component ablation on KITTI 00--10.}
MAGC combines inherited memory augmentation (MA) with our
ground conditioning (GC). Checks indicate active components.
We report mean ATE (m); lower is better.}
\label{tab:ablation_factors}
\vspace{-4pt}
\end{table}

With MAGC active, SA and CL individually yield ATEs of
8.32\,m and 8.18\,m, respectively, compared with 8.36\,m
for MAGC alone. Enabling both reduces ATE to 7.30\,m,
a 12.7\% improvement. The larger joint gain suggests that SA's local alignment gains are better retained when CL preserves selected inter-chunk relations
through shared corrections, preventing independent chunk
updates from altering these relations during global refinement. With MA alone, jointly enabling SA and CL reduces ATE
by 1.7\% (13.65 to 13.42\,m), compared with 12.7\%
with MAGC. This further supports their complementary
use with ground conditioning.


\Needspace{12\baselineskip}
\section{Conclusion}
\label{sec:conclusion}

We presented Embodied-VGGT, a framework for long sequence 3D reconstruction that exploits geometric and temporal cues from embodied acquisition without external camera poses. MAGC uses predicted ground geometry to regularize memory adaptation and establish relative scales, while SA and CL jointly establish and selectively preserve reliable inter-chunk relationships during global refinement. Experiments demonstrate substantial improvements in trajectory accuracy across all five evaluated driving benchmarks, with point-cloud evaluation on KITTI further confirming improved reconstruction quality. These findings highlight the value of embodied geometric information for correcting global drift while retaining consistent local structure in long sequence 3D reconstruction.


\clearpage

\section*{LLM Usage Statement}

OpenAI ChatGPT was used to assist with editing and
refining author-produced manuscript text, while
OpenAI Codex assisted with reviewing and refining
author-developed code and implementation.
ChatGPT was also used to obtain reviewer-style
feedback on the manuscript.

No LLMs or other AI tools were used for related work
and citations, experimental design, execution or analysis,
figure preparation, data creation or annotation,
or the formulation of research ideas and conclusions.

We take full responsibility for all submitted content,
regardless of any LLM or AI-tool use.


{
    \small
    \bibliographystyle{ieeenat_fullname}
    \bibliography{main}

@inproceedings{Wang_2025_CVPR,
  author = {Wang, Jianyuan and Chen, Minghao and Karaev, Nikita and Vedaldi, Andrea and Rupprecht, Christian and Novotny, David},
  title = {{VGGT}: Visual Geometry Grounded Transformer},
  booktitle = {Proceedings of the IEEE/CVF Conference on Computer Vision and Pattern Recognition (CVPR)},
  year = {2025},
  pages = {5294--5306}
}

@inproceedings{deng2025vggtlongchunkitloop,
  author = {Deng, Kai and Ti, Zexin and Xu, Jiawei and Yang, Jian and Xie, Jin},
  title = {{VGGT-Long}: Chunk it, Loop it, Align it -- Pushing {VGGT}'s Limits on Kilometer-scale Long {RGB} Sequences},
  booktitle = {IEEE International Conference on Robotics and Automation (ICRA)},
  year = {2026}
}

@inproceedings{Xie_2026_CVPR,
  author = {Xie, Tao and Yang, Peishan and Jin, Yudong and Cai, Yingfeng and Yin, Wei and Ren, Weiqiang and Zhang, Qian and Hua, Wei and Peng, Sida and Guo, Xiaoyang and Zhou, Xiaowei},
  title = {{Scal3R}: Scalable Test-Time Training for Large-Scale {3D} Reconstruction},
  booktitle = {Proceedings of the IEEE/CVF Conference on Computer Vision and Pattern Recognition (CVPR)},
  year = {2026},
  pages = {21760--21771}
}

@inproceedings{zhang2026vggt,
  author = {Zhang, Wei and Wu, Yihang and Li, Songhua and Wang, Qi},
  title = {{VGGT-Align}: Bridging Local Reconstruction and Global Consistency for Long-Sequence {3D} Reconstruction},
  booktitle = {Proceedings of the 34th ACM International Conference on Multimedia (ACM MM)},
  year = {2026},
  doi = {10.1145/3767308.3836543},
  eprint = {2608.15260},
  archivePrefix = {arXiv}
}

@inproceedings{Zhang_2026_CVPR,
  author = {Zhang, Fengyi and Zhang, Tianjun and Khosoussi, Kasra and Zhang, Zheng and Huang, Zi and Luo, Yadan},
  title = {{TALO}: Pushing {3D} Vision Foundation Models Towards Globally Consistent Online Reconstruction},
  booktitle = {Proceedings of the IEEE/CVF Conference on Computer Vision and Pattern Recognition (CVPR)},
  year = {2026},
  pages = {21870--21879}
}

@inproceedings{Wang_2024_CVPR,
  author = {Wang, Shuzhe and Leroy, Vincent and Cabon, Yohann and Chidlovskii, Boris and Revaud, Jerome},
  title = {{DUSt3R}: Geometric {3D} Vision Made Easy},
  booktitle = {Proceedings of the IEEE/CVF Conference on Computer Vision and Pattern Recognition (CVPR)},
  year = {2024},
  pages = {20697--20709}
}

@inproceedings{Leroy_2024_ECCV,
  author = {Leroy, Vincent and Cabon, Yohann and Revaud, Jerome},
  title = {Grounding Image Matching in {3D} with {MASt3R}},
  booktitle = {Proceedings of the European Conference on Computer Vision (ECCV)},
  year = {2024},
  doi = {10.1007/978-3-031-73220-1_5},
  pages = {71--91}
}

@inproceedings{Wang_CUT3R_2025_CVPR,
  author = {Wang, Qianqian and Zhang, Yifei and Holynski, Aleksander and Efros, Alexei A. and Kanazawa, Angjoo},
  title = {Continuous {3D} Perception Model with Persistent State},
  booktitle = {Proceedings of the IEEE/CVF Conference on Computer Vision and Pattern Recognition (CVPR)},
  year = {2025},
  pages = {10510--10522}
}

@inproceedings{Murai_2025_CVPR,
  author = {Murai, Riku and Dexheimer, Eric and Davison, Andrew J.},
  title = {{MASt3R-SLAM}: Real-Time Dense {SLAM} with {3D} Reconstruction Priors},
  booktitle = {Proceedings of the IEEE/CVF Conference on Computer Vision and Pattern Recognition (CVPR)},
  year = {2025},
  pages = {16695--16705}
}

@inproceedings{Shen_2026_ICLR,
  author = {Shen, You and Zhang, Zhipeng and Qu, Yansong and Zheng, Xiawu and Ji, Jiayi and Zhang, Shengchuan and Cao, Liujuan},
  title = {{FastVGGT}: Fast Visual Geometry Transformer},
  booktitle = {International Conference on Learning Representations (ICLR)},
  year = {2026}
}

@inproceedings{Schonberger_2016_CVPR,
  author = {Sch{\"o}nberger, Johannes L. and Frahm, Jan-Michael},
  title = {Structure-from-Motion Revisited},
  booktitle = {Proceedings of the IEEE Conference on Computer Vision and Pattern Recognition (CVPR)},
  year = {2016},
  pages = {4104--4113}
}

@inproceedings{Teed_2021_NeurIPS,
  author = {Teed, Zachary and Deng, Jia},
  title = {{DROID-SLAM}: Deep Visual {SLAM} for Monocular, Stereo, and {RGB-D} Cameras},
  booktitle = {Advances in Neural Information Processing Systems (NeurIPS)},
  year = {2021},
  pages = {16558--16569},
  volume = {34}
}

@inproceedings{Teed_2023_NeurIPS,
  author = {Teed, Zachary and Lipson, Lahav and Deng, Jia},
  title = {Deep Patch Visual Odometry},
  booktitle = {Advances in Neural Information Processing Systems (NeurIPS)},
  year = {2023},
  volume = {36},
  pages = {39033--39051},
  doi = {10.52202/075280-1696}
}

@inproceedings{Song_2014_CVPR,
  author = {Song, Shiyu and Chandraker, Manmohan},
  title = {Robust Scale Estimation in Real-Time Monocular {SFM} for Autonomous Driving},
  booktitle = {Proceedings of the IEEE Conference on Computer Vision and Pattern Recognition (CVPR)},
  year = {2014},
  pages = {1566--1573}
}

@article{Song_2016_TPAMI,
  author = {Song, Shiyu and Chandraker, Manmohan and Guest, Clark C.},
  title = {High Accuracy Monocular {SFM} and Scale Correction for Autonomous Driving},
  journal = {IEEE Transactions on Pattern Analysis and Machine Intelligence},
  year = {2016},
  volume = {38},
  number = {4},
  pages = {730--743},
  doi = {10.1109/TPAMI.2015.2469274}
}

@article{Zhou_2020_TITS,
  author = {Zhou, Dingfu and Dai, Yuchao and Li, Hongdong},
  title = {Ground-Plane-Based Absolute Scale Estimation for Monocular Visual Odometry},
  journal = {IEEE Transactions on Intelligent Transportation Systems},
  year = {2020},
  volume = {21},
  number = {2},
  pages = {791--802},
  doi = {10.1109/TITS.2019.2900330}
}

@inproceedings{Zhou_2016_ECCV,
  author = {Zhou, Qian-Yi and Park, Jaesik and Koltun, Vladlen},
  title = {Fast Global Registration},
  booktitle = {Proceedings of the European Conference on Computer Vision (ECCV)},
  year = {2016},
  pages = {766--782},
  doi = {10.1007/978-3-319-46475-6_47}
}

@article{Yang_2021_TRO,
  author = {Yang, Heng and Shi, Jingnan and Carlone, Luca},
  title = {{TEASER}: Fast and Certifiable Point Cloud Registration},
  journal = {IEEE Transactions on Robotics},
  year = {2021},
  volume = {37},
  number = {2},
  pages = {314--333},
  doi = {10.1109/TRO.2020.3033695}
}

@inproceedings{Qin_2022_CVPR,
  author = {Qin, Zheng and Yu, Hao and Wang, Changjian and Guo, Yulan and Peng, Yuxing and Xu, Kai},
  title = {Geometric Transformer for Fast and Robust Point Cloud Registration},
  booktitle = {Proceedings of the IEEE/CVF Conference on Computer Vision and Pattern Recognition (CVPR)},
  year = {2022},
  pages = {11143--11152}
}

@inproceedings{Huang_2021_CVPR,
  author = {Huang, Shengyu and Gojcic, Zan and Usvyatsov, Mikhail and Wieser, Andreas and Schindler, Konrad},
  title = {{Predator}: Registration of {3D} Point Clouds with Low Overlap},
  booktitle = {Proceedings of the IEEE/CVF Conference on Computer Vision and Pattern Recognition (CVPR)},
  year = {2021},
  pages = {4267--4276}
}

@inproceedings{Bai_2021_CVPR,
  author = {Bai, Xuyang and Luo, Zixin and Zhou, Lei and Chen, Hongkai and Li, Lei and Hu, Zeyu and Fu, Hongbo and Tai, Chiew-Lan},
  title = {{PointDSC}: Robust Point Cloud Registration Using Deep Spatial Consistency},
  booktitle = {Proceedings of the IEEE/CVF Conference on Computer Vision and Pattern Recognition (CVPR)},
  year = {2021},
  pages = {15859--15869}
}

@inproceedings{Landrieu_2018_CVPR,
  author = {Landrieu, Loic and Simonovsky, Martin},
  title = {Large-Scale Point Cloud Semantic Segmentation with Superpoint Graphs},
  booktitle = {Proceedings of the IEEE Conference on Computer Vision and Pattern Recognition (CVPR)},
  year = {2018},
  pages = {4558--4567}
}

@inproceedings{Landrieu_2019_CVPR,
  author = {Landrieu, Loic and Boussaha, Mohamed},
  title = {Point Cloud Oversegmentation with Graph-Structured Deep Metric Learning},
  booktitle = {Proceedings of the IEEE/CVF Conference on Computer Vision and Pattern Recognition (CVPR)},
  year = {2019},
  pages = {7440--7449}
}

@inproceedings{Yew_2020_CVPR,
  author = {Yew, Zi Jian and Lee, Gim Hee},
  title = {{RPM-Net}: Robust Point Matching Using Learned Features},
  booktitle = {Proceedings of the IEEE/CVF Conference on Computer Vision and Pattern Recognition (CVPR)},
  year = {2020},
  pages = {11824--11833}
}

@inproceedings{Sarlin_2020_CVPR,
  author = {Sarlin, Paul-Edouard and DeTone, Daniel and Malisiewicz, Tomasz and Rabinovich, Andrew},
  title = {{SuperGlue}: Learning Feature Matching with Graph Neural Networks},
  booktitle = {Proceedings of the IEEE/CVF Conference on Computer Vision and Pattern Recognition (CVPR)},
  year = {2020},
  pages = {4938--4947}
}

@article{Umeyama_1991_TPAMI,
  author = {Umeyama, Shinji},
  title = {Least-Squares Estimation of Transformation Parameters Between Two Point Patterns},
  journal = {IEEE Transactions on Pattern Analysis and Machine Intelligence},
  year = {1991},
  volume = {13},
  number = {4},
  pages = {376--380},
  doi = {10.1109/34.88573}
}

@article{Fischler_1981_CACM,
  author = {Fischler, Martin A. and Bolles, Robert C.},
  title = {Random Sample Consensus: A Paradigm for Model Fitting with Applications to Image Analysis and Automated Cartography},
  journal = {Communications of the ACM},
  year = {1981},
  volume = {24},
  number = {6},
  pages = {381--395},
  doi = {10.1145/358669.358692}
}

@article{Mur_Artal_2015_TRO,
  author = {Mur-Artal, Raul and Montiel, J. M. M. and Tardos, Juan D.},
  title = {{ORB-SLAM}: A Versatile and Accurate Monocular {SLAM} System},
  journal = {IEEE Transactions on Robotics},
  year = {2015},
  volume = {31},
  number = {5},
  pages = {1147--1163},
  doi = {10.1109/TRO.2015.2463671}
}

@article{Campos_2021_TRO,
  author = {Campos, Carlos and Elvira, Richard and Gomez Rodriguez, Juan J. and Montiel, Jose M. M. and Tardos, Juan D.},
  title = {{ORB-SLAM3}: An Accurate Open-Source Library for Visual, Visual-Inertial, and Multimap {SLAM}},
  journal = {IEEE Transactions on Robotics},
  year = {2021},
  volume = {37},
  number = {6},
  pages = {1874--1890},
  doi = {10.1109/TRO.2021.3075644}
}

@inproceedings{Engel_2014_ECCV,
  author = {Engel, Jakob and Schops, Thomas and Cremers, Daniel},
  title = {{LSD-SLAM}: Large-Scale Direct Monocular {SLAM}},
  booktitle = {Proceedings of the European Conference on Computer Vision (ECCV)},
  year = {2014},
  pages = {834--849},
  doi = {10.1007/978-3-319-10605-2_54}
}

@inproceedings{Whelan_2015_RSS,
  author = {Whelan, Thomas and Leutenegger, Stefan and Salas-Moreno, Renato and Glocker, Ben and Davison, Andrew J.},
  title = {{ElasticFusion}: Dense {SLAM} Without a Pose Graph},
  booktitle = {Robotics: Science and Systems (RSS)},
  year = {2015},
  doi = {10.15607/RSS.2015.XI.001}
}

@article{Dai_2017_TOG,
  author = {Dai, Angela and Nie{\ss}ner, Matthias and Zollh{\"o}fer, Michael and Izadi, Shahram and Theobalt, Christian},
  title = {{BundleFusion}: Real-Time Globally Consistent {3D} Reconstruction Using On-the-Fly Surface Reintegration},
  journal = {ACM Transactions on Graphics},
  year = {2017},
  volume = {36},
  number = {3},
  pages = {24:1--24:18},
  doi = {10.1145/3054739}
}

@inproceedings{Kuemmerle_2011_ICRA,
  author = {K{\"u}mmerle, Rainer and Grisetti, Giorgio and Strasdat, Hauke and Konolige, Kurt and Burgard, Wolfram},
  title = {{g2o}: A General Framework for Graph Optimization},
  booktitle = {IEEE International Conference on Robotics and Automation (ICRA)},
  year = {2011},
  pages = {3607--3613},
  doi = {10.1109/ICRA.2011.5979949}
}

@inproceedings{Suenderhauf_2012_IROS,
  author = {S{\"u}nderhauf, Niko and Protzel, Peter},
  title = {Switchable Constraints for Robust Pose Graph {SLAM}},
  booktitle = {IEEE/RSJ International Conference on Intelligent Robots and Systems (IROS)},
  pages = {1879--1884},
  year = {2012},
  doi = {10.1109/IROS.2012.6385590}
}

@inproceedings{Grisetti_2010_ICRA,
  author = {Grisetti, Giorgio and K{\"u}mmerle, Rainer and Stachniss, Cyrill and Frese, Udo and Hertzberg, Christoph},
  title = {Hierarchical Optimization on Manifolds for Online {2D} and {3D} Mapping},
  booktitle = {IEEE International Conference on Robotics and Automation (ICRA)},
  year = {2010},
  pages = {273--278},
  doi = {10.1109/ROBOT.2010.5509407}
}

@inproceedings{Geiger2012CVPR,
  author = {Geiger, Andreas and Lenz, Philip and Urtasun, Raquel},
  title = {Are We Ready for Autonomous Driving? The {KITTI} Vision Benchmark Suite},
  booktitle = {Proceedings of the IEEE Conference on Computer Vision and Pattern Recognition (CVPR)},
  pages = {3354--3361},
  year = {2012}
}

@inproceedings{Gaidon_2016_CVPR,
  author = {Gaidon, Adrien and Wang, Qiao and Cabon, Yohann and Vig, Eleonora},
  title = {Virtual Worlds as Proxy for Multi-Object Tracking Analysis},
  booktitle = {Proceedings of the IEEE Conference on Computer Vision and Pattern Recognition (CVPR)},
  year = {2016},
  pages = {4340--4349}
}

@inproceedings{Caesar_2020_CVPR,
  author = {Caesar, Holger and Bankiti, Varun and Lang, Alex H. and Vora, Sourabh and Liong, Venice Erin and Xu, Qiang and Krishnan, Anush and Pan, Yu and Baldan, Giancarlo and Beijbom, Oscar},
  title = {{nuScenes}: A Multimodal Dataset for Autonomous Driving},
  booktitle = {Proceedings of the IEEE/CVF Conference on Computer Vision and Pattern Recognition (CVPR)},
  year = {2020},
  pages = {11621--11631}
}

@InProceedings{Cheng_MERG3R_2026_CVPR,
  author = {Cheng, Leo Kaixuan and Shaikh, Abdus
            and Liang, Ruofan and Wu, Zhijie
            and Guan, Yushi and Vijaykumar, Nandita},
  title = {{MERG3R}: A Divide-and-Conquer Approach
           to Large-Scale Neural Visual Geometry},
  booktitle = {Proceedings of the IEEE/CVF Conference
               on Computer Vision and Pattern
               Recognition (CVPR)},
  year = {2026},
  pages = {28969--28978}
}

@InProceedings{Elflein_2026_CVPR,
  author = {Elflein, Sven and Li, Ruilong
            and Agostinho, Sergio and Gojcic, Zan
            and Leal-Taixe, Laura and Zhou, Qunjie
            and Osep, Aljosa},
title = {{VGG-T}$^{3}$: Offline Feed-Forward
         {3D} Reconstruction at Scale},
  booktitle = {Proceedings of the IEEE/CVF Conference
               on Computer Vision and Pattern
               Recognition (CVPR)},
  year = {2026},
  pages = {36464--36474}
}

@InProceedings{Ding_LASER_2026_CVPR,
  author = {Ding, Tianye and Xie, Yiming
            and Liang, Yiqing and Chatterjee, Moitreya
            and Miraldo, Pedro and Jiang, Huaizu},
  title = {{LASER}: Layer-wise Scale Alignment
           for Training-Free Streaming
           {4D} Reconstruction},
  booktitle = {Proceedings of the IEEE/CVF Conference
               on Computer Vision and Pattern
               Recognition (CVPR)},
  year = {2026},
  pages = {36433--36443}
}

@article{Achanta_2012_PAMI,
  author = {Achanta, Radhakrishna and Shaji, Appu and Smith, Kevin and Lucchi, Aurelien and Fua, Pascal and S{\"u}sstrunk, Sabine},
  title = {{SLIC} Superpixels Compared to State-of-the-Art Superpixel Methods},
  journal = {IEEE Transactions on Pattern Analysis and Machine Intelligence},
  year = {2012},
  volume = {34},
  number = {11},
  pages = {2274--2282},
  doi = {10.1109/TPAMI.2012.120}
}

@inproceedings{Chen_TTT3R_2026_ICLR,
  author = {Chen, Xingyu and Chen, Yue and Xiu, Yuliang and Geiger, Andreas and Chen, Anpei},
  title = {{TTT3R}: {3D} Reconstruction as Test-Time Training},
  booktitle = {International Conference on Learning Representations (ICLR)},
  year = {2026}
}

@InProceedings{Cheng_LongStream_2026_CVPR,
  author = {Cheng, Chong and Chen, Xianda and Xie, Tao and Yin, Wei and Ren, Weiqiang and Zhang, Qian and Guo, Xiaoyang and Wang, Hao},
  title = {{LongStream}: Long-Sequence Streaming Autoregressive Visual Geometry},
  booktitle = {Proceedings of the IEEE/CVF Conference on Computer Vision and Pattern Recognition (CVPR)},
  year = {2026},
  pages = {272--283}
}

@inproceedings{Yang_Fast3R_2025_CVPR,
  author = {Yang, Jianing and Sax, Alexander and Liang, Kevin J. and Henaff, Mikael and Tang, Hao and Cao, Ang and Chai, Joyce and Meier, Franziska and Feiszli, Matt},
  title = {{Fast3R}: Towards {3D} Reconstruction of 1000+ Images in One Forward Pass},
  booktitle = {Proceedings of the IEEE/CVF Conference on Computer Vision and Pattern Recognition (CVPR)},
  year = {2025},
  pages = {21924--21935}
}

@inproceedings{Lan_STream3R_2026_ICLR,
  author = {Lan, Yushi and Luo, Yihang and Hong, Fangzhou and Zhou, Shangchen and Chen, Honghua and Lyu, Zhaoyang and Yang, Shuai and Dai, Bo and Loy, Chen Change and Pan, Xingang},
  title = {{STream3R}: Scalable Sequential {3D} Reconstruction with Causal Transformer},
  booktitle = {International Conference on Learning Representations (ICLR)},
  year = {2026}
}

@inproceedings{Lee_SwiftVGGT_2025_arxiv,
  author = {Lee, Jungho and Lee, Minhyeok and Yang, Sunghun and Kang, Minseok and Lee, Sangyoun},
  title = {{SwiftVGGT}: A Scalable Visual Geometry Grounded Transformer for Large-Scale Scenes},
  booktitle = {Proceedings of the IEEE/CVF Conference on Computer Vision and Pattern Recognition (CVPR) Findings},
  year = {2026}
}

@inproceedings{Sun_2020_CVPR,
  author = {Sun, Pei and Kretzschmar, Henrik and Dotiwalla, Xerxes
    and Chouard, Aurelien and Patnaik, Vijaysai and Tsui, Paul
    and Guo, James and Zhou, Yin and Chai, Yuning and Caine, Benjamin
    and Vasudevan, Vijay and Han, Wei and Ngiam, Jiquan
    and Zhao, Hang and Timofeev, Aleksei and Ettinger, Scott
    and Krivokon, Maxim and Gao, Amy and Joshi, Aditya
    and Zhang, Yu and Shlens, Jonathon and Chen, Zhifeng
    and Anguelov, Dragomir},
  title = {Scalability in Perception for Autonomous Driving:
    Waymo Open Dataset},
  booktitle = {Proceedings of the IEEE/CVF Conference on
    Computer Vision and Pattern Recognition (CVPR)},
  pages = {2446--2454},
  year = {2020}
}

@misc{Cabon_2020_VKITTI2,
  author = {Cabon, Yohann and Murray, Naila and Humenberger, Martin},
  title = {Virtual {KITTI} 2},
  year = {2020},
  note = {arXiv:2001.10773}
}
}


\clearpage
\begingroup
\setcounter{page}{1}
\setcounter{section}{0}
\setcounter{equation}{24}
\setcounter{table}{6}
\setcounter{figure}{3}
\renewcommand{\thesection}{\Alph{section}}
\raggedbottom

\maketitlesupplementary

\setlength{\parskip}{0pt plus .5pt}
\setlength{\parindent}{1pc}
\setlength{\abovedisplayskip}{6pt plus 2pt minus 2pt}
\setlength{\belowdisplayskip}{6pt plus 2pt minus 2pt}
\setlength{\abovedisplayshortskip}{3pt plus 1pt}
\setlength{\belowdisplayshortskip}{4pt plus 1pt}
\setlength{\jot}{3pt}
\setlength{\intextsep}{15pt plus 1pt minus 1pt}
\setlength{\textfloatsep}{\intextsep}
\setlength{\dbltextfloatsep}{\intextsep}
\setlength{\floatsep}{\intextsep}
\setlength{\dblfloatsep}{\intextsep}
\setcounter{topnumber}{3}
\setcounter{dbltopnumber}{3}
\renewcommand{\topfraction}{.92}
\renewcommand{\dbltopfraction}{.95}
\renewcommand{\textfraction}{.05}
\renewcommand{\dblfloatpagefraction}{.75}
\captionsetup[table]{position=bottom,skip=6pt}
\captionsetup[figure]{position=bottom,skip=6pt}


\section{Method Details}
\label{sec:suppl_method_details}

\subsection{Ground References and MAGC Memory Update}
\label{sec:magc_scale_derivation_supp}

\noindent\textbf{Ground references.}
For the initial fit, we sample the lower 45\% of every third frame
at four-pixel intervals along both image axes. We keep points above
mean frame confidence whose vertical offsets from their camera centers fall
between the 5th and 50th percentiles along the first camera's up axis.
Confidence-weighted PCA with four rounds of residual trimming must
retain at least 1,000 points and yield a normal within $8^\circ$ of
this axis. In Eqs.~(2)--(4) of the main paper, $h_{if}^{0}$ is the
signed distance from the initial camera center at frame $f$ to the
initial plane of chunk $i$. Shape at frame $f$ uses $h_{if}^{0}$,
whereas motion from $u$ to $v$ uses $h_{iu}^{0}$. Both references
stay fixed during adaptation. A common rescaling of the chunk geometry multiplies
each distance or displacement and its reference height by the same
factor, leaving the normalized relation unchanged. After refinement,
we refit the plane using every frame. The median camera-to-plane
distance $H_i$ supplies the inter-chunk scale in Eq.~(9) of the main paper.

\noindent\textbf{Reference construction.}
In overlapping chunks, we sample matching pixels on a $16\times16$
grid. A frame with at least eight valid points above the mean confidence
in each prediction supplies up to 128 point pairs. Motion references
use shared frame pairs 5 or 15 frames apart. Shape disagreement
is the absolute difference in log normalized distances. For motion, we
divide the norm of the displacement difference by the mean of the two
displacement norms, floored at 0.1. Shape and motion disagreements up to
0.03 and 0.05, respectively, enter the preservation set $\mathcal K$.
Values above those thresholds and up to 0.15 and 0.20 enter the
correction set $\mathcal R$.
Both sets keep their initial memberships and consensus targets during
the update. In Eq.~(5) of the main paper, each chunk prediction is
compared with the same target, defined using prediction confidence
and ground-fit quality. For shape, a group contains point-pair
observations from one shared frame. For motion, it contains the
displacement observation for one shared frame pair. Averaging within
each group before averaging across groups gives each group equal
weight even when frames supply different numbers of point pairs.

\noindent\textbf{Memory update.}
A fixed hash of frame IDs (seed 0) assigns 40\% of buckets to fitting,
20\% to checking the proposed update, and the remainder to neither
step. A physical frame keeps its assignment wherever it appears, and
the endpoints of a motion observation must share an assignment.
With pretrained parameters fixed, we compute one gradient for the
three runtime memory tensors in Table~\ref{tab:suppl_impl_settings}
($K=1$). We test the three listed step multipliers along this same
gradient direction, largest first. We accept the first step whose
held-out loss is finite and at most 0.002 above the pre-update loss.
If either split lacks support or no candidate satisfies this check,
we decode with the original memory weights.

\noindent\textbf{Relative scale.}
Let $a_i$ convert physical distances to chunk-$i$ units, so
$H_i\simeq a_i h_i^{\mathrm{phys}}$ for physical platform--ground
distance $h_i^{\mathrm{phys}}$. Locally stable heights give
\begin{equation}
s_{i\leftarrow j}^{\mathrm{MAGC}}
=\frac{H_i}{H_j}\simeq\frac{a_i}{a_j},
\qquad h_i^{\mathrm{phys}}\simeq h_j^{\mathrm{phys}}.
\end{equation}
Each chunk reuses $H_i$ on all valid grounded edges, so for $j>i$
\begin{equation}
\label{eq:magc_transport_supp}
\prod_{k=i}^{j-1}s_{k\leftarrow k+1}^{\mathrm{MAGC}}
=\prod_{k=i}^{j-1}\frac{H_k}{H_{k+1}}
=\frac{H_i}{H_j}.
\end{equation}
These ratios determine the chunk scales up to a common factor.

\newcounter{embImplTable}
\renewcommand{\theembImplTable}{A\arabic{embImplTable}}
\par\addvspace{\intextsep}
\noindent\begin{minipage}{\columnwidth}
\captionsetup{type=table,hypcap=false}
\centering\small
\setlength{\tabcolsep}{2pt}
\renewcommand{\arraystretch}{1.14}
\begin{tabular*}{\columnwidth}{@{\extracolsep{\fill}}ll@{}}
\toprule
\textbf{Implementation setting} & \textbf{Value} \\
\midrule
\multicolumn{2}{@{}l}{\textit{MAGC memory update}} \\
Adapted global block (zero-based) & 20 ($w_0,w_1,w_2$) \\
Motion / preservation weights ($\beta_m,\beta_{\rm keep}$) & $0.5$ / $1.0$ \\
Huber threshold & $0.10$ \\
Relative step / norm cap & $10^{-4}$ / $10^{-3}$ \\
Step coefficient cap / gradient floor & $10^{5}$ / $10^{-8}$ \\
Step multipliers, in trial order & $1,\ 1/2,\ 1/4$ \\
\midrule
\multicolumn{2}{@{}l}{\textit{SA translation and CL optimization}} \\
Radial Huber coefficient (SA) & $1.345$ \\
Initial PGO damping (CL) & $10^{-6}$ \\
Maximum PGO iterations (CL) & 30 \\
\bottomrule
\end{tabular*}
\par
\refstepcounter{embImplTable}\label{tab:suppl_impl_settings}
\caption*{\textbf{Table~\theembImplTable. Settings in the released implementation.}
The relative step and norm cap are fractions of the current weight
norm, applied before scaling by $\lambda$.}
\end{minipage}\par
\addvspace{\intextsep}

\subsection{Shared Spatial Partition and Support Classification}
\label{sec:sa_partition_supp}

For each shared frame $f$, SA assigns a common region label to
corresponding pixels in chunks $i$ and $j$. We express each point
in the camera coordinates of its own reconstruction. In these native length
units, the shared cue is
\begin{equation}
\label{eq:sp_shared_view}
\overline{\mathbf p}_{f,p}^{\mathrm{cam}}=\tfrac12\!\left[
(\mathbf C_f^i)^{-1}\!\cdot\mathbf y_p+
(\mathbf C_f^j)^{-1}\!\cdot\mathbf x_p\right].
\end{equation}
From this averaged map we compute unit surface normals and divide
depth by the median absolute depth over valid pixels (floored at
$10^{-4}$). The SLIC-inspired partition~[\textcolor{cvprblue}{1}]
starts from an $8\times8$ grid and runs five iterations. Its
assignment distance combines squared differences in RGB, geometry
(normalized depth and normals), and pixel position, with geometry weight
$0.10$ and spatial compactness $0.65$. This averaging
serves only to assign regions. Translation residuals use the original correspondences after applying
the estimated relative scale and rotation.

For each reconstruction, we fit a local plane by confidence-weighted
PCA using lower-image points within the chunk ground band whose
confidence exceeds the frame mean. We require
at least 100 points and check each normal against its chunk normal
and the two normals against each other after rotation ($8^\circ$
angular tolerance). For chunk $i$, we set the band width
$\beta_{i,f}=\operatorname{clip}(3\sigma_{i,f},0.015H_i,0.20H_i)$,
where $\sigma_{i,f}$ is $1.4826$ times the median absolute deviation
of the absolute point-to-plane distances, floored at $10^{-6}$.
For accepted planes, the absolute
point-to-plane distances $\delta_p^i,\delta_p^j$ determine the label
\begin{equation}
\label{eq:sp_support_class}
\ell_p=\begin{cases}
\mathrm G,&\delta_p^i\le\beta_{i,f},\
           \delta_p^j\le\beta_{j,f},\\
\mathrm N,&\delta_p^i>2\beta_{i,f},\
           \delta_p^j>2\beta_{j,f},\\
\mathrm U,&\text{otherwise}.
\end{cases}
\end{equation}
If a plane fails the quality checks, all matches in that frame
receive label $\mathrm U$.

Each region contributes a ground or non-ground translation estimate if
the corresponding category has at least eight positive-weight
matches. Each category must contribute at least three such regions
across at least three distinct frames. We aggregate region estimates
within frames and then across frames using equal-weight geometric
medians, and average the two category estimates. If either category
lacks sufficient support, we estimate translation
with a Huber loss at fixed scale and rotation. If a chunk height or ground band is unavailable, we retain
the original visual similarity transform for that edge.

\subsection{Cross-Window Selection and Group Corrections}
\label{sec:cl_interpretation_supp}

For each triplet of consecutive ordinary chunks $q=(A,B,C)$, we test both overlaps
$(A,B)$ and $(B,C)$. Each must contain at least 12 shared frames.
We sample six frames uniformly per overlap. A frame contributes
to overlap-scale inconsistency (OS) if its $32\times32$ sampling
grid yields at least 256 valid correspondences and 64 valid point
pairs. The distance between the two points must exceed $0.02D_B$
in both reconstructions, where $D_B$ is the median of the middle chunk's
per-frame median camera-to-point distances over supported frames.
Each overlap must supply at least four such frames.
We take the $90$th percentile of the absolute log distance ratio
in the initial global frame, first over pairs and then over frames. We use the larger
of the two overlap scores.

For cross-window motion inconsistency (XT), span $h\in\{1,2,4,8\}$
links the $h$th frame from the end of $A$ to the $h$th frame of $C$.
Both endpoints must occur in $B$ in that order. We require at least
three valid spans, including $h=1$. XT is the $75$th percentile
of the discrepancy between the outer and middle chunks' relative
displacements, normalized by the frame separation and the middle
chunk's median camera-step length, floored at $0.001D_B$.

Among candidates that pass the support checks and have finite,
nonnegative diagnostics, we rank OS and XT separately within
each sequence. Tied values receive their average rank. For $M$
valid candidates, a one-based rank $r$ becomes $(r-0.5)/M$.
For normalized ranks $u,v$, we scan in ascending order of
$\max(u,v)+(u+v)/8$, up to the default budget in
Eq.~\eqref{eq:suppl_candidate_scan_budget}. We break ties by the
mean of the two normalized ranks and then by the triplet's starting index.
We skip triplets that share a selected chunk or contain both
endpoints of a loop edge. We stop at the budget even when a candidate
is skipped. Each ungrouped chunk forms a singleton.

With loop edges, we jointly optimize one left $\mathrm{Sim}(3)$
correction per group using the adjacent and loop constraints.
Sharing a correction preserves relative transforms within each
triplet. The group containing the first chunk is fixed.
Without loop edges, we compose the adjacent transforms directly.


\section{Dataset Details}
\label{sec:suppl_datasets}

We evaluate Embodied-VGGT on three driving benchmarks
in the main paper: KITTI Odometry~[\textcolor{cvprblue}{16}],
Virtual KITTI 1~[\textcolor{cvprblue}{15}], and
nuScenes mini~[\textcolor{cvprblue}{4}].
These datasets cover diverse driving environments,
trajectory lengths, and visual conditions.

\noindent\textbf{KITTI Odometry}~[\textcolor{cvprblue}{16}]
is a real-world driving benchmark comprising 11 sequences
(00--10) collected in urban, residential, and highway
environments. Sequence lengths range from 271 to 4,661
frames, covering trajectories of up to several kilometers
and providing challenging conditions for long-sequence
reconstruction and camera pose estimation.

\noindent\textbf{Virtual KITTI 1}~[\textcolor{cvprblue}{15}]
is a synthetic driving dataset featuring different weather
and illumination conditions. We evaluate Scene 20 under
six conditions: Clone, Fog, Morning, Overcast, Rain,
and Sunset. Each sequence contains 837 frames, with
identical scene geometry and camera trajectories across
the six conditions.

{
\raggedright
\noindent\textbf{nuScenes mini}~[\textcolor{cvprblue}{4}]
is a real-world autonomous driving dataset with diverse
urban scenes and traffic conditions. We evaluate ten
scenes using front-facing RGB images (CAM\_FRONT):
\texttt{scene-0061}, \texttt{scene-0103},
\texttt{scene-0553}, \texttt{scene-0655},
\texttt{scene-0757}, \texttt{scene-0796},
\texttt{scene-0916}, \texttt{scene-1077},
\texttt{scene-1094}, and \texttt{scene-1100}.
The sequences contain 224--240 frames.
\par
}

We additionally evaluate Embodied-VGGT on two driving
datasets in the supplementary material:
Waymo Open Dataset~[\textcolor{cvprblue}{33}] and
Virtual KITTI 2~[\textcolor{cvprblue}{3}].
These datasets extend the evaluation to additional
real-world and synthetic driving environments.

\noindent\textbf{Waymo Open Dataset}~[\textcolor{cvprblue}{33}]
is a large-scale real-world autonomous driving dataset
featuring diverse road layouts, traffic conditions,
and camera motions. We evaluate nine driving sequences
using RGB images from the front-facing camera.

\noindent\textbf{Virtual KITTI 2}~[\textcolor{cvprblue}{3}]
is a synthetic driving dataset with improved visual
realism and diverse environmental conditions.
We evaluate five sequences:
\texttt{Scene01\_clone},
\texttt{Scene02\_30-deg-right},
\texttt{Scene06\_30-deg-left},
\texttt{Scene18\_30-deg-left}, and
\texttt{Scene20\_fog}.
These sequences cover different driving scenes,
camera viewpoints, and visual conditions.
We use Virtual KITTI~2 for in-domain evaluation.


\section{Evaluation Details}
\label{sec:suppl_evaluation}

We provide further details on the evaluation protocols
for camera pose accuracy and 3D reconstruction quality.

\noindent\textbf{Pose metrics.}
We evaluate camera pose accuracy using the
\texttt{evo} toolkit (v1.38.0).
We align predicted camera trajectories with ground truth
using a positive-scale Sim(3) transformation to account
for differences in global coordinate systems and scale.

We use Absolute Trajectory Error (ATE) as the primary
pose evaluation metric across all five benchmarks:
KITTI Odometry, Virtual KITTI 1, nuScenes mini,
Waymo Open Dataset, and Virtual KITTI 2.
ATE measures the root-mean-square translation error
between aligned camera centers and ground-truth
positions, reported in meters.

For KITTI Odometry, we additionally report Relative
Translation Error (RTE, m/100\,m) and Relative
Rotation Error (RRE, deg/100\,m).
We evaluate these metrics over 100\,m trajectory
segments. For each starting frame, we select the
first subsequent frame whose accumulated ground-truth
travel distance reaches at least 100\,m.
We then compute the relative pose errors between
the corresponding frame pairs.

We compute dataset-level mean errors over the
designated evaluation sequences of each benchmark.
We report complete-run means only for methods
successfully evaluated on all required sequences.
When a trajectory contains no valid 100\,m segment,
the corresponding relative errors are marked as
unavailable.

\noindent\textbf{Reconstruction metrics.}
For each KITTI Odometry sequence, we construct a reference
point cloud from the Velodyne scans using the provided
calibration and ground-truth camera poses.
We transform each scan into the rectified camera coordinate
system and retain points within the image boundaries and
a camera-frame $z$-depth range of 1--80\,m.
The retained points are then accumulated in a common
sequence coordinate system.

We estimate a single $\mathrm{Sim}(3)$ transformation
from the predicted and ground-truth camera centers over
the full sequence and apply it to the predicted point cloud.
Before depth filtering, we convert predicted camera-frame
$z$ depths to meters using the estimated scale.
For chunk-local predictions from Embodied-VGGT, this
conversion also accounts for the chunk-to-sequence scale.
We retain predicted points within the same metric depth
range and image boundaries, together with the
model-provided valid-point masks.
For methods providing confidence scores, we additionally
retain points whose confidence exceeds $0.75$ times the
mean predicted confidence.

We independently voxel-downsample the predicted and
reference point clouds using a voxel size of 0.10\,m,
representing each occupied voxel by the centroid of its
points. We then refine the sequence-level point-cloud
alignment using point-to-point ICP before computing
reconstruction metrics. No framewise or local point-cloud
registration is performed.
Ground-truth scans and poses are used only for evaluation.

Let $\mathcal{P}$ and $\mathcal{G}$ denote the resulting
predicted and reference point clouds, respectively.
Let $d(\mathbf{x}\!\to\!\mathcal{S})
= \min_{\mathbf{y}\in\mathcal{S}}\|\mathbf{x}-\mathbf{y}\|_2$
denote the distance from a point $\mathbf{x}$ to its
nearest neighbor in the point cloud $\mathcal{S}$.
Accuracy (Acc.) and completeness (Comp.) average this
distance from prediction to reference and from reference
to prediction, respectively:
\begin{align}
\mathrm{Acc.}
&= \frac{1}{|\mathcal{P}|}
\sum_{\mathbf{p}\in\mathcal{P}}
d(\mathbf{p}\!\to\!\mathcal{G}),
\label{eq:suppl_recon_acc}\\
\mathrm{Comp.}
&= \frac{1}{|\mathcal{G}|}
\sum_{\mathbf{g}\in\mathcal{G}}
d(\mathbf{g}\!\to\!\mathcal{P}).
\label{eq:suppl_recon_comp}
\end{align}
The Chamfer distance (CD) is their arithmetic mean:
\begin{equation}
\mathrm{CD}
= \frac{\mathrm{Acc.}+\mathrm{Comp.}}{2}.
\label{eq:suppl_recon_cd}
\end{equation}
All nearest-neighbor distances contribute to these means
without distance truncation.
We apply the same evaluation protocol to all compared
methods. We compute all three metrics in meters for each
sequence and report their arithmetic means across KITTI
Odometry sequences 00--10.


\begin{table*}[!t]
\centering
\small
\setlength{\tabcolsep}{1pt}
\renewcommand{\arraystretch}{1.2}
\begin{tabular*}{\linewidth}
{@{\extracolsep{\fill}}l*{9}{c}>{\centering\arraybackslash}p{32pt}@{}}
\toprule
\textbf{Methods} & \textbf{S1} & \textbf{S2}
& \textbf{S3} & \textbf{S4} & \textbf{S5}
& \textbf{S6} & \textbf{S7} & \textbf{S8}
& \textbf{S9} & \textbf{Mean} \\
\midrule
DROID-SLAM~[\textcolor{cvprblue}{35}]
& 2.64 & \textbf{0.08} & \textbf{0.20} & 9.53
& 12.42 & 5.45 & 3.20 & 23.43 & \textbf{0.09} & 6.34 \\
MASt3R-SLAM~[\textcolor{cvprblue}{26}]
& 4.43 & 0.36 & 1.24 & 5.31 & 6.52
& 2.63 & 3.23 & 8.15 & 1.45 & 3.70 \\
\addlinespace[2pt]
CUT3R~[\textcolor{cvprblue}{39}]
& 7.70 & 2.32 & 4.17 & 27.53 & 16.89
& 5.48 & 14.16 & 11.21 & 1.78 & 10.14 \\
Fast3R~[\textcolor{cvprblue}{44}]
& 46.07 & 11.74 & 47.56 & 102.20 & 80.61
& 23.71 & 78.04 & 34.22 & 19.57 & 49.30 \\
STream3R~[\textcolor{cvprblue}{20}]
& {\color{black!55}\textit{OOM}}
& {\color{black!55}\textit{OOM}}
& {\color{black!55}\textit{OOM}}
& {\color{black!55}\textit{OOM}}
& {\color{black!55}\textit{OOM}}
& {\color{black!55}\textit{OOM}}
& {\color{black!55}\textit{OOM}}
& {\color{black!55}\textit{OOM}}
& {\color{black!55}\textit{OOM}}
& -- \\
TTT3R~[\textcolor{cvprblue}{6}]
& 2.47 & 1.40 & 2.19 & 5.96 & 7.73
& 1.63 & 2.80 & 4.33 & 1.84 & 3.37 \\
\addlinespace[2pt]
VGGT-Long~[\textcolor{cvprblue}{10}]
& \textbf{1.78} & 0.23 & 0.72 & 3.11 & \textbf{3.22}
& 1.39 & \textbf{1.70} & 3.47 & 0.17 & 1.75 \\
Scal3R~[\textcolor{cvprblue}{42}]
& 4.17 & 0.33 & 0.46 & \textbf{0.89} & 5.16
& 2.12 & 2.84 & 6.93 & 0.28 & 2.58 \\
VGGT-Align~[\textcolor{cvprblue}{47}]
& 3.90 & 0.36 & 0.49 & 1.87 & 3.58
& \textbf{1.11} & 2.46 & 2.08 & 2.27 & 2.01 \\
\midrule
\textbf{Embodied-VGGT (Ours)}
& 3.32 & 0.17 & 0.28 & 0.92 & 3.82
& 1.63 & 2.33 & \textbf{2.04} & 0.15 & \textbf{1.63} \\
\bottomrule
\end{tabular*}
\par
\caption{\textbf{Camera pose accuracy on the Waymo Open Dataset.}
Per-sequence absolute trajectory error (ATE, m) and mean ATE
over nine driving sequences. Lower is better. Bold indicates the
lowest error in each column. -- denotes no complete-run mean;
\textit{OOM}: out of memory.
S1--S9: 163453191, 183829460, 315615587, 346181117,
371159869, 405841035, 460417311, 520018670, 610454533.}
\label{tab:suppl_pose_waymo}
\end{table*}

\begin{table*}[!b]
\centering
\small
\setlength{\tabcolsep}{0pt}
\renewcommand{\arraystretch}{1.2}
\begin{tabular}{@{}
>{\raggedright\arraybackslash}m{0.28\linewidth}
*{5}{>{\centering\arraybackslash}m{0.125\linewidth}}
>{\centering\arraybackslash}m{0.095\linewidth}
@{}}
\toprule
\textbf{Methods}
& \shortstack[c]{\textbf{Scene01}\\\textbf{Clone}}
& \shortstack[c]{\textbf{Scene02}\\\textbf{30-deg-right}}
& \shortstack[c]{\textbf{Scene06}\\\textbf{30-deg-left}}
& \shortstack[c]{\textbf{Scene18}\\\textbf{30-deg-left}}
& \shortstack[c]{\textbf{Scene20}\\\textbf{Fog}}
& \textbf{Mean} \\
\midrule
DROID-SLAM~[\textcolor{cvprblue}{35}]
& 1.05 & 0.08 & \textbf{0.04} & 0.50 & 8.38 & 2.01 \\
MASt3R-SLAM~[\textcolor{cvprblue}{26}]
& 78.94 & 30.88 & 2.51 & 23.69
& {\color{black!55}\textit{PF}} & -- \\
\addlinespace[2pt]
CUT3R~[\textcolor{cvprblue}{39}]
& 59.10 & 33.26 & 0.39 & 34.54 & 117.45 & 48.95 \\
Fast3R~[\textcolor{cvprblue}{44}]
& {\color{black!55}\textit{PF}} & 33.72 & 11.34
& 70.48 & {\color{black!55}\textit{PF}} & -- \\
STream3R~[\textcolor{cvprblue}{20}]
& 60.70 & 9.45 & 0.19 & 53.33
& {\color{black!55}\textit{OOM}} & -- \\
TTT3R~[\textcolor{cvprblue}{6}]
& 28.62 & 15.72 & 0.49 & 8.08 & 75.52 & 25.69 \\
\addlinespace[2pt]
VGGT-Long~[\textcolor{cvprblue}{10}]
& 2.20 & 0.18 & 0.20 & 0.51 & 6.27 & 1.87 \\
Scal3R~[\textcolor{cvprblue}{42}]
& 0.40 & 0.10 & 0.06 & 0.38 & 3.28 & 0.85 \\
VGGT-Align~[\textcolor{cvprblue}{47}]
& 3.51 & 1.16 & 3.00 & 1.49 & 5.27 & 2.89 \\
\midrule
\textbf{Embodied-VGGT (Ours)}
& \textbf{0.30} & \textbf{0.07} & 0.05
& \textbf{0.10} & \textbf{1.15} & \textbf{0.33} \\
\bottomrule
\end{tabular}
\par
\caption{\textbf{Camera pose accuracy on Virtual KITTI 2.}
Per-sequence absolute trajectory error (ATE, m) and mean ATE
over five selected sequences. Lower is better. Bold indicates the
lowest error in each column. -- denotes no complete-run mean;
\textit{OOM}: out of memory; \textit{PF}: pose-estimation failure.}
\label{tab:suppl_pose_vkitti2}
\end{table*}

\section{Experimental Details}
\label{sec:suppl_experimental_details}

\subsection{Experimental Settings}
\label{sec:suppl_experimental_settings}

All evaluations use the full set of frames per sequence.
Across all datasets, we set the chunk size to 60 frames
and the overlap to 30 frames for Embodied-VGGT,
VGGT-Long~[\textcolor{cvprblue}{10}],
Scal3R~[\textcolor{cvprblue}{42}],
TALO~[\textcolor{cvprblue}{46}],
VGGT-Align~[\textcolor{cvprblue}{47}], and
SwiftVGGT~[\textcolor{cvprblue}{23}].
For the remaining baselines~[\textcolor{cvprblue}{6, 20, 26, 30, 35, 39, 44}],
we follow their official inference protocols.

\subsection{Memory Management Strategy}
\label{sec:suppl_memory_management}

We conduct our experiments on a workstation running
Ubuntu 22.04.5 LTS, equipped with an Intel Core Ultra 9
285K CPU, 192\,GiB of RAM, and two NVIDIA RTX 5880 Ada
Generation GPUs with 48\,GiB of VRAM each.
Each inference run uses a single GPU.

To manage memory consumption on long sequences, we process
one sequence at a time. Within each sequence, we schedule
independent chunk forward passes while limiting the number
of chunks resident on the GPU at any given time.
Following the offloading strategy of
VGGT-Long~[\textcolor{cvprblue}{10}], we transfer each
completed chunk's predictions to CPU memory or disk and
release its intermediate GPU tensors, making room for
subsequent chunks. Once all chunk predictions are available,
we perform the alignment and optimization stages
described in the main paper.


\begin{figure*}[!t]
\raggedright
\includegraphics[width=\linewidth]{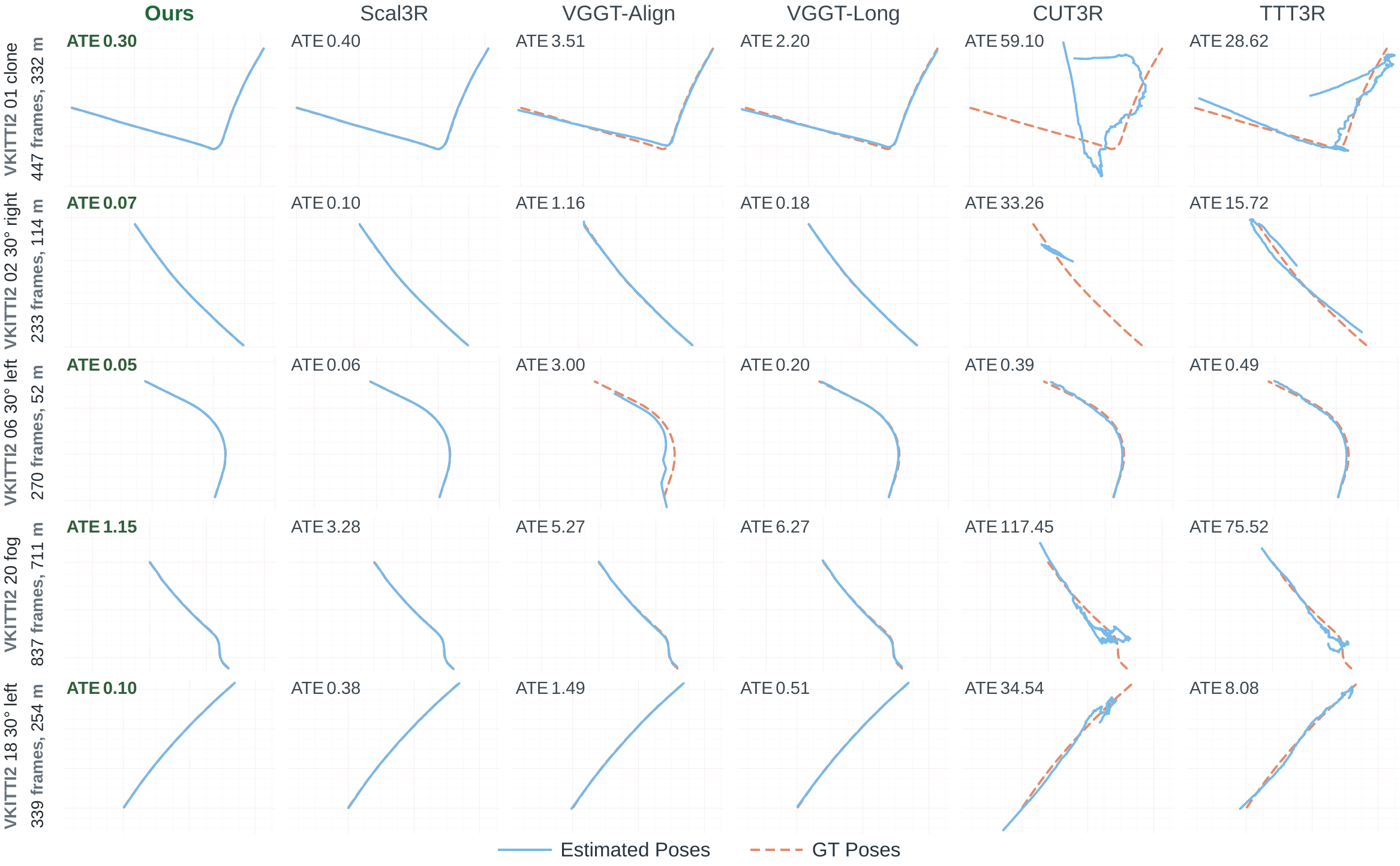}
\caption{\textbf{Camera trajectory comparisons on Virtual KITTI 2.}
Estimated camera trajectories and ground truth on five selected
sequences for Embodied-VGGT and five representative baselines.
ATE values (m) are displayed in each panel.}
\label{fig:suppl_vkitti2_trajectories}
\end{figure*}

\begin{table*}[!b]
\centering
\small
\setlength{\tabcolsep}{1pt}
\renewcommand{\arraystretch}{1.12}
\begin{tabular*}{\linewidth}
{@{\extracolsep{\fill}}l*{12}{c}@{}}
\toprule
\textbf{Variant}
& \textbf{00} & \textbf{01} & \textbf{02} & \textbf{03}
& \textbf{04} & \textbf{05} & \textbf{06} & \textbf{07}
& \textbf{08} & \textbf{09} & \textbf{10} & \textbf{Mean} \\
\midrule
VGGT-Align~[\textcolor{cvprblue}{47}] & 7.68 & 65.26 & 46.99 & 5.04 & 1.73 & 4.92 & 4.23 & 4.47 & 44.10 & 23.24 & 11.77 & 19.95 \\
Scal3R~[\textcolor{cvprblue}{42}] & 4.18 & 50.47 & 37.77 & 5.22 & \textbf{1.05} & 3.84 & 3.37 & 1.36 & 25.29 & \textbf{12.43} & 5.96 & 13.72 \\
VGGT-Align + MA & 12.37 & 76.82 & 32.78 & 4.69 & 7.45 & 3.09 & 4.93 & 2.01 & 31.43 & 17.35 & 6.78 & 18.15 \\
\midrule
\textbf{MAGC (Ours)} & \textbf{3.51} & \textbf{19.79} & \textbf{20.57} & \textbf{1.57} & 1.15 & \textbf{2.92} & \textbf{1.36} & \textbf{1.33} & \textbf{20.94} & 13.88 & \textbf{4.92} & \textbf{8.36} \\
\bottomrule
\end{tabular*}
\par
\caption{\textbf{Effectiveness of MAGC on KITTI Odometry.}
Per-sequence and mean ATE (m) over sequences 00--10.
MA denotes memory augmentation.
VGGT-Align + MA uses Scal3R for chunk inference while
retaining the remaining VGGT-Align pipeline and settings.
Lower is better. Best results are highlighted in bold.}
\label{tab:suppl_magc_components}
\end{table*}
\begin{figure*}[!t]
\centering
\includegraphics[width=.91\linewidth]{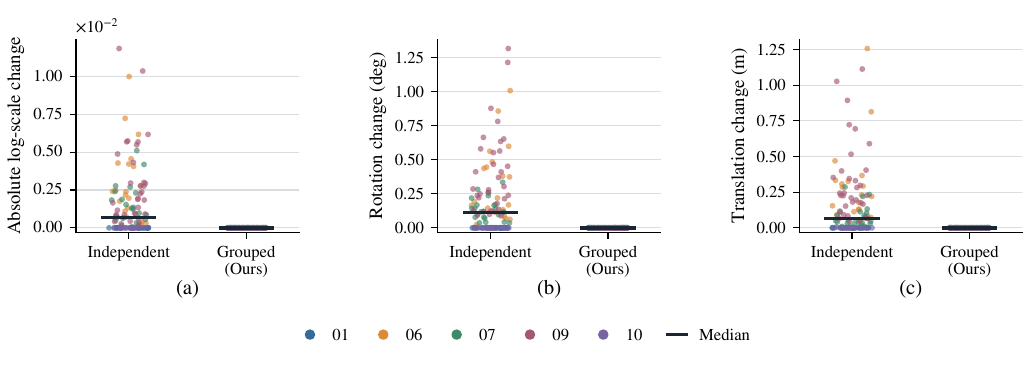}
\caption{\textbf{Preservation of within-group relative geometry.}
For each of the 123 within-group chunk pairs, we compare its
relative transform before and after PGO. Panels show changes in
(a) log scale, (b) rotation, and (c) translation under independent
and grouped corrections. Colors denote KITTI sequences and black
bars mark medians. Translation changes are expressed in meters
using fixed scales fitted to ground-truth camera centers.}
\label{fig:suppl_group_relative_changes}
\end{figure*}

\section{Additional Experiments}
\label{sec:suppl_additional_experiments}

\subsection{Waymo \& Virtual KITTI 2}
\label{sec:suppl_waymo_vkitti2}

We provide additional camera pose evaluations on the Waymo
Open Dataset and Virtual KITTI 2.
Table~\ref{tab:suppl_pose_waymo} and
Table~\ref{tab:suppl_pose_vkitti2} report per-sequence and mean ATE.
Embodied-VGGT achieves mean ATEs of 1.63\,m on Waymo
and 0.33\,m on Virtual KITTI 2.

On Waymo, Embodied-VGGT reduces mean ATE from 2.58\,m
for our reproduced Scal3R baseline to 1.63\,m, a 36.8\%
reduction, with improvements on eight of the nine sequences.
Since our framework builds on the memory-augmented predictions
of Scal3R, these results support the effectiveness of the
additional geometric modeling in our framework on this benchmark.

Both Embodied-VGGT and our reproduced Scal3R baseline use
60-frame chunks with a 30-frame overlap and are evaluated
on the same nine sequences using the protocol in Sec.~C.
Our reproduced Scal3R result is 2.58\,m, whereas the original
study~[\textcolor{cvprblue}{42}] reports 1.52\,m.
Table~\ref{tab:suppl_pose_waymo} reports our reproduced results, and the relative
improvement above is computed against this reproduced baseline.

On the five selected Virtual KITTI 2 sequences, Embodied-VGGT achieves a mean ATE of 0.33 m and the lowest ATE on four sequences (Table~\ref{tab:suppl_pose_vkitti2}). It improves on Scal3R on all five, with the largest absolute reduction on Scene20 Fog, from 3.28 m to 1.15 m (Fig.~\ref{fig:suppl_vkitti2_trajectories}). On Scene06, DROID-SLAM achieves 0.04 m and Embodied-VGGT achieves 0.05 m.

\subsection{Additional Experiments and Results}
\label{sec:suppl_additional_diagnostics}

We conduct six experiments.
First, we compare MAGC with VGGT-Align, Scal3R, and VGGT-Align + MA.
Second, we compare CL-score, chronological, and random group
selection with matched counts.
Third, we compare independent and grouped optimization.
Fourth, we reduce ground-reference geometric support.
Fifth, we perturb reference heights.
Sixth, we vary candidate search budgets for ATE and group coverage.

All experiments follow Sec.~\ref{sec:suppl_evaluation} and
Sec.~\ref{sec:suppl_experimental_details}.
The MAGC comparison uses KITTI 00--10. The other five experiments use full sequences
01, 06, 07, 09, and 10. Randomized runs use seeds 17, 29, and 41,
averaged within each sequence and then across sequences.

\newsavebox{\supplGroupSelectionTable}
\begin{lrbox}{\supplGroupSelectionTable}
\begin{minipage}{\columnwidth}
\centering
\small
\renewcommand{\arraystretch}{1.12}
\begin{tabular*}{\linewidth}
{@{\extracolsep{\fill}}lc@{}}
\toprule
\textbf{Grouping} & \textbf{Mean} \\
\midrule
\textbf{CL score (Ours)} & \textbf{7.90} \\
Fixed order & 8.58 \\
Random valid & $8.21 \pm 0.25$ \\
\bottomrule
\end{tabular*}
\par
\captionsetup{hypcap=false}
\captionof{table}{\textbf{Effect of group selection on the five KITTI sequences.}
All strategies use the same candidate pool and number of selected
groups per sequence. For Random valid, we report the mean and sample
standard deviation of the three seed-specific dataset means.
Lower is better.}
\label{tab:suppl_group_selection}
\end{minipage}
\end{lrbox}

\begin{table*}[!b]
\centering
\small
\setlength{\tabcolsep}{1.1pt}
\renewcommand{\arraystretch}{1.12}
\begin{tabular*}{\linewidth}
{@{\extracolsep{\fill}}lcccccc@{}}
\toprule
\multirow{2}{*}{\textbf{Optimization}}
& \multicolumn{3}{c}{\textbf{Camera poses}}
& \multicolumn{3}{c}{\textbf{Reconstruction}} \\
\cmidrule(lr){2-4}\cmidrule(l){5-7}
& \textbf{ATE} & \textbf{RTE} & \textbf{RRE}
& \textbf{Acc.} & \textbf{Comp.} & \textbf{CD} \\
\midrule
Before optimization & 11.38 & 4.94 & 2.24 & 4.31 & 1.54 & 2.93 \\
Independent PGO & 8.22 & 4.86 & 2.12 & 3.80 & 1.19 & 2.49 \\
\textbf{Grouped PGO (Ours)}
& \textbf{7.90} & \textbf{4.73} & \textbf{2.07}
& \textbf{3.76} & \textbf{1.15} & \textbf{2.46} \\
\bottomrule
\end{tabular*}
\par
\caption{\textbf{Effect of the PGO parameterization.}
Mean errors on KITTI sequences 01, 06, 07, 09, and 10.
Independent and grouped PGO start from the same initial estimates
and use the same front-end predictions, graph constraints, and
30-iteration limit. ATE, Acc., Comp., and CD are in meters;
RTE is in m/100\,m and RRE in deg/100\,m.
Reconstruction metrics follow Sec.~\ref{sec:suppl_evaluation}. Lower is better.
}
\label{tab:suppl_grouped_pgo}
\end{table*}

\subsubsection{MAGC Component Comparison}
\label{sec:suppl_magc_components}

To assess the effectiveness of MAGC, we compare it with
VGGT-Align, Scal3R, and VGGT-Align + MA
on KITTI 00--10 (Table~\ref{tab:suppl_magc_components}).
For VGGT-Align + MA, we replace VGGT with Scal3R for chunk
inference, keeping the remaining VGGT-Align pipeline and
settings unchanged.
MAGC incorporates ground-referenced geometry into memory
adaptation and estimates relative scales from the refined
predictions.
It achieves the lowest mean ATE of 8.36\,m, compared with
19.95\,m for VGGT-Align, 13.72\,m for Scal3R, and 18.15\,m
for their direct combination, and obtains the lowest ATE
on nine of the eleven sequences.
Compared with VGGT-Align + MA, MAGC improves all eleven
sequences and reduces mean ATE by 53.9\%, supporting its
effectiveness beyond directly combining the existing components.

\subsubsection{Group Selection}
\label{sec:suppl_group_selection}

Table~\ref{tab:suppl_group_selection} compares CL-score,
chronological (Fixed order), and Random valid selection with matched
group counts and identical candidate pools, non-overlap and
loop-endpoint rules, and optimization settings.
Mean ATE is 7.90\,m, 8.58\,m, and $8.21\pm0.25$\,m, respectively.
The sample standard deviation is computed over the three
seed-specific dataset means.
CL-score reduces mean ATE by 0.68\,m over Fixed order and
0.31\,m over Random valid.

\begin{figure*}[!t]
\centering
\includegraphics[width=0.75\linewidth]{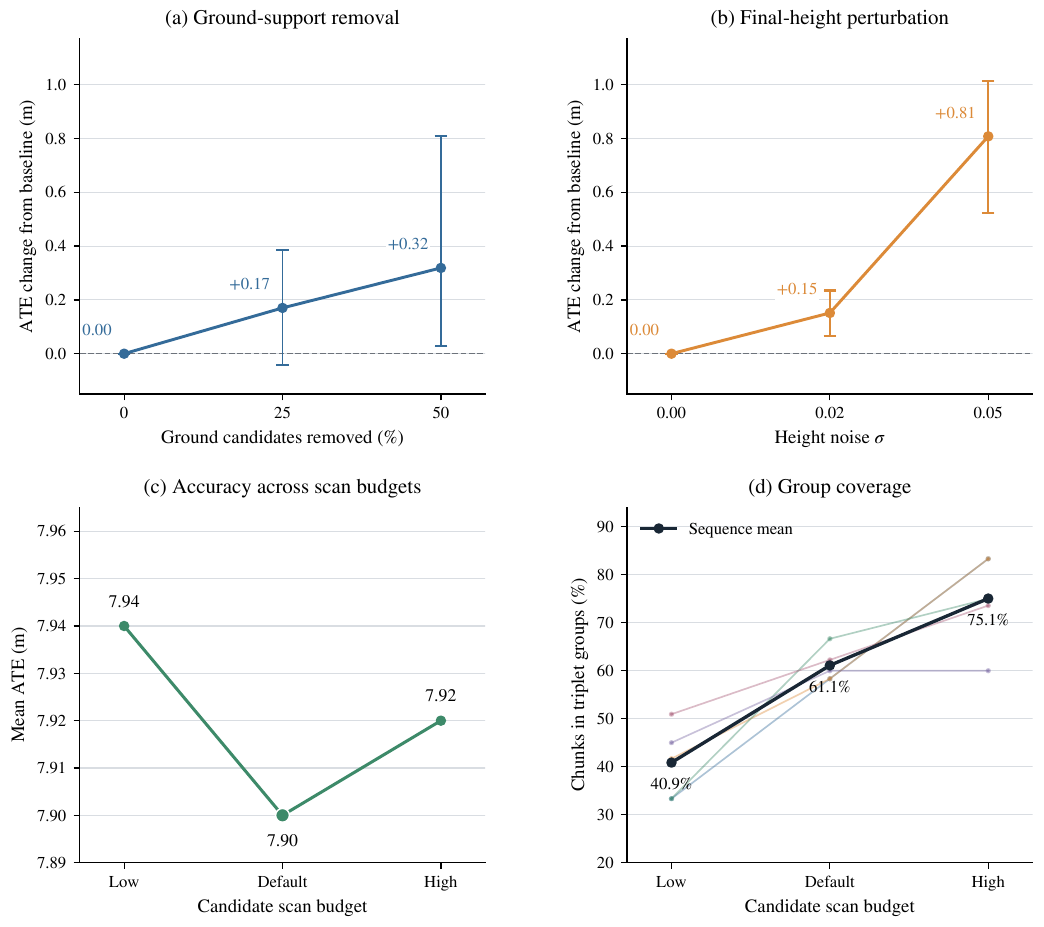}
\caption{\textbf{Sensitivity to ground references and candidate
search budget.} On the five-sequence KITTI subset, (a) and (b)
show the change in mean ATE relative to the unperturbed baseline
(7.90\,m) under ground-support removal and final-height
perturbation, respectively. Error bars in (a,b) denote 95\%
bootstrap confidence intervals over sequences after averaging
the three seeds within each sequence. Panels (c) and (d) show
mean ATE and the percentage of chunks assigned to triplet groups
as the scan budget increases. In (d), thin colored lines denote
individual sequences and the black line denotes their mean.
}
\label{fig:suppl_sensitivity}
\end{figure*}

\par\addvspace{\intextsep}
\noindent\usebox{\supplGroupSelectionTable}\par
\addvspace{\intextsep}

\subsubsection{Grouped Optimization}
\label{sec:suppl_grouped_optimization}

Table~\ref{tab:suppl_grouped_pgo} compares independent and grouped PGO using identical
front-end predictions, graph constraints, initialization, and a
30-iteration limit. On the five-sequence subset, grouped PGO
reduces mean ATE from 8.22 to 7.90\,m and mean CD from 2.49
to 2.46\,m. Mean RTE and RRE also decrease. Figure~\ref{fig:suppl_group_relative_changes} verifies
the preservation property of the shared correction in
Eq.~(23) of the main paper: across the 123 within-group chunk
pairs, the largest translation change after grouped PGO is
below 0.2\,mm.

\subsubsection{Geometric Support Removal}
\label{sec:suppl_geometric_support}

With RGB inputs and confidence scores fixed, we mask contiguous
horizontal bands of XYZ candidates for ground-reference estimation.
Removing 25\% and 50\% of the ground candidates increases mean ATE
by 0.17\,m and 0.32\,m, respectively
(Fig.~\ref{fig:suppl_sensitivity}(a)).

\subsubsection{Reference Height Perturbation}
\label{sec:suppl_reference_height}

For independent $\epsilon_i\sim\mathcal N(0,\sigma^2)$ with
$\sigma\in\{0,0.02,0.05\}$, we perturb final ground-reference heights as
\begin{equation}
\begin{aligned}
H_i' &= H_i\exp(\epsilon_i),\\
s_{i\leftarrow j}' &= s_{i\leftarrow j}\exp(\epsilon_i-\epsilon_j).
\end{aligned}
\label{eq:suppl_height_perturbation}
\end{equation}

Each chunk uses the same perturbed height on all incident
height-derived edges, so intermediate height factors still
cancel along a chain as in Eq.~(26). This experiment measures
trajectory sensitivity to errors in the per-chunk ground references.

Relative to the unperturbed baseline, mean ATE increases by
0.15\,m at $\sigma=0.02$ and by 0.81\,m at $\sigma=0.05$
(Fig.~\ref{fig:suppl_sensitivity}(b)).

\subsubsection{Candidate Search Budget}
\label{sec:suppl_candidate_budget}

For $N$ chunks and $M$ valid candidates, set
$K_0=\min\{M,\lfloor(N-2)/2\rfloor\}>0$.
We vary only the number of ranked candidates scanned:
\begin{equation}
K_{\mathrm{scan}}=
\begin{cases}
\max\{1,\lfloor K_0/2\rfloor\},&\text{Low},\\
K_0,&\text{Default},\\
\min\{M,\lceil3K_0/2\rceil\},&\text{High}.
\end{cases}
\label{eq:suppl_candidate_scan_budget}
\end{equation}

Figure~\ref{fig:suppl_sensitivity}(c,d) shows how the scan budget affects trajectory accuracy and group coverage. For $K$ selected triplets among $N$ chunks,
coverage is $3K/N$. Increasing the budget from Low to Default
to High raises mean coverage from 40.86\% to 61.12\% and 75.05\%,
respectively. The corresponding mean ATEs are 7.94, 7.90, and
7.92\,m, spanning only 0.04\,m across the three budgets.

\clearpage
\endgroup

\end{document}